\documentclass[10pt,twocolumn,letterpaper]{article}

\usepackage{3dv}
\usepackage{times}
\usepackage{epsfig}
\usepackage{graphicx}
\usepackage{amsmath}
\usepackage{amssymb}

\usepackage{subfig}
\usepackage{array}
\usepackage{textcomp}
\usepackage[table]{xcolor}
\usepackage{algorithm}
\usepackage[noend]{algpseudocode}
\def\BState{\State\hskip-\ALG@thistlm}

\usepackage[pagebackref=true,breaklinks=true,letterpaper=true,colorlinks,bookmarks=false]{hyperref}

\threedvfinalcopy 

\def\threedvPaperID{120} 

\begin{document}

\title{DUGMA: Dynamic Uncertainty-Based Gaussian Mixture Alignment}

\author{Can Pu \\
	can.pu@ed.ac.uk\\
\and
Nanbo Li \\
	nanbo.li@ed.ac.uk\\
\and
Radim Tylecek \\
	rtylecek@inf.ed.ac.uk\\
\and
Robert B. Fisher \\ 
	rbf@inf.ed.ac.uk\\
School of Informatics, University of Edinburgh
}

\maketitle

\begin{abstract}
	Accurately registering point clouds from a cheap low-resolution sensor is a challenging task. Existing rigid registration methods failed to use the physical 3D uncertainty distribution of each point from a real sensor in the dynamic alignment process. It is mainly because the uncertainty model for a point is static and invariant and it is hard to describe the change of these physical uncertainty models in different views. Additionally, the existing Gaussian mixture alignment architecture cannot efficiently implement these dynamic changes. 
	
	This paper proposes a simple architecture combining error estimation from sample covariances and dynamic global probability alignment using the convolution of uncertainty-based Gaussian Mixture Models (GMM). Firstly, we propose an efficient way to describe the change of each 3D uncertainty model, which represents the structure of the point cloud better. Unlike the invariant GMM (representing a fixed point cloud) in traditional Gaussian mixture alignment, we use two uncertainty-based GMMs that change and interact with each other in each iteration. In order to have a wider basin of convergence than other local algorithms, we design a more robust energy function by convolving efficiently the two GMMs over the whole 3D space. 
	
	Tens of thousands of trials have been conducted on hundreds of models from multiple datasets to demonstrate the proposed method's superior performance compared with the current state-of-the-art methods. All the materials including our code are available from \url{https://github.com/Canpu999/DUGMA}.     
\end{abstract}


\section{Introduction}
\begin{figure}[h!]
	\centering
	\subfloat[Before registration]{\includegraphics[width = 4cm,height=3cm]{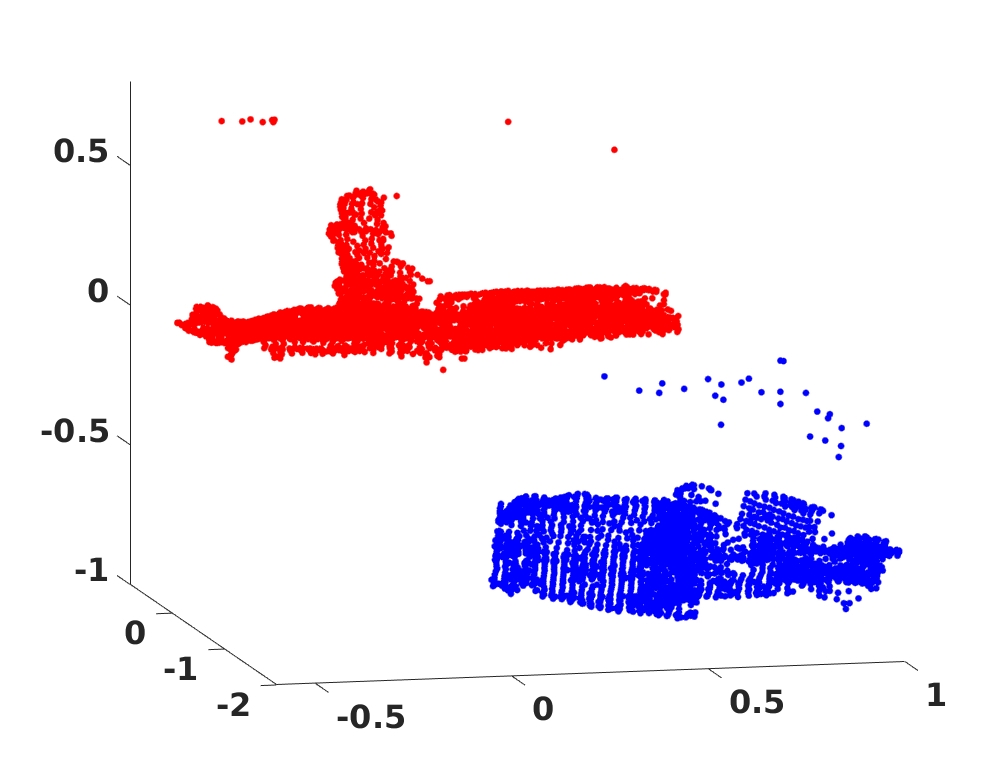}}
	\subfloat[After registration]{\includegraphics[width = 4cm,height=3cm]{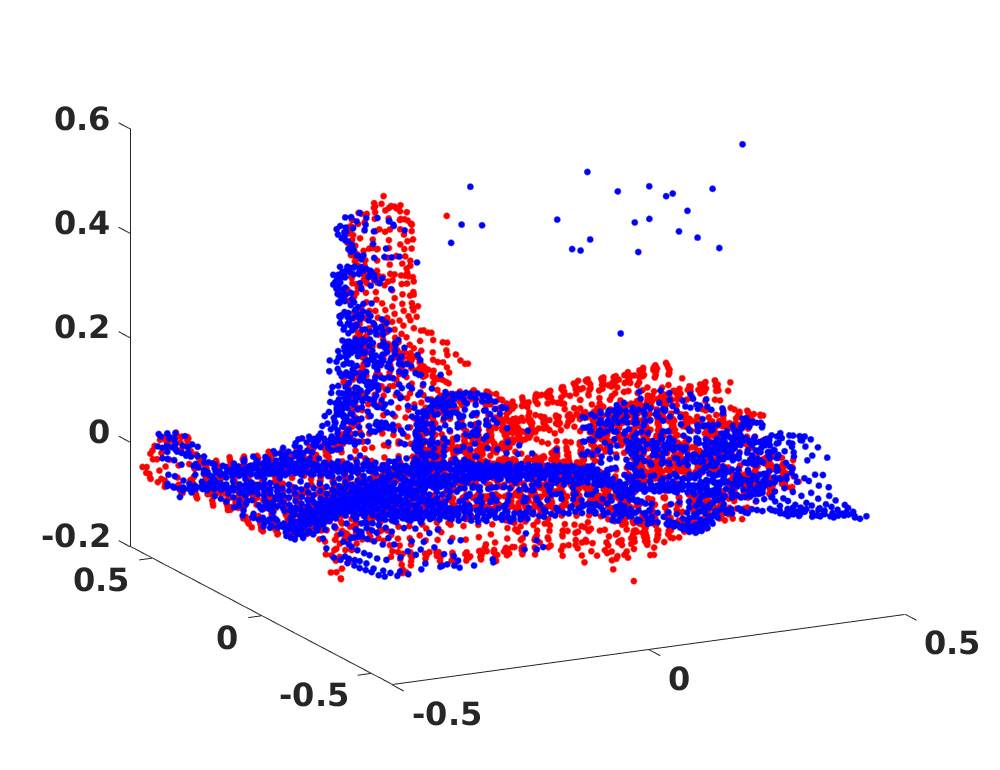}} \\
	\subfloat[After registration (texture mapped)]{\includegraphics[width = 7cm,height=5cm]{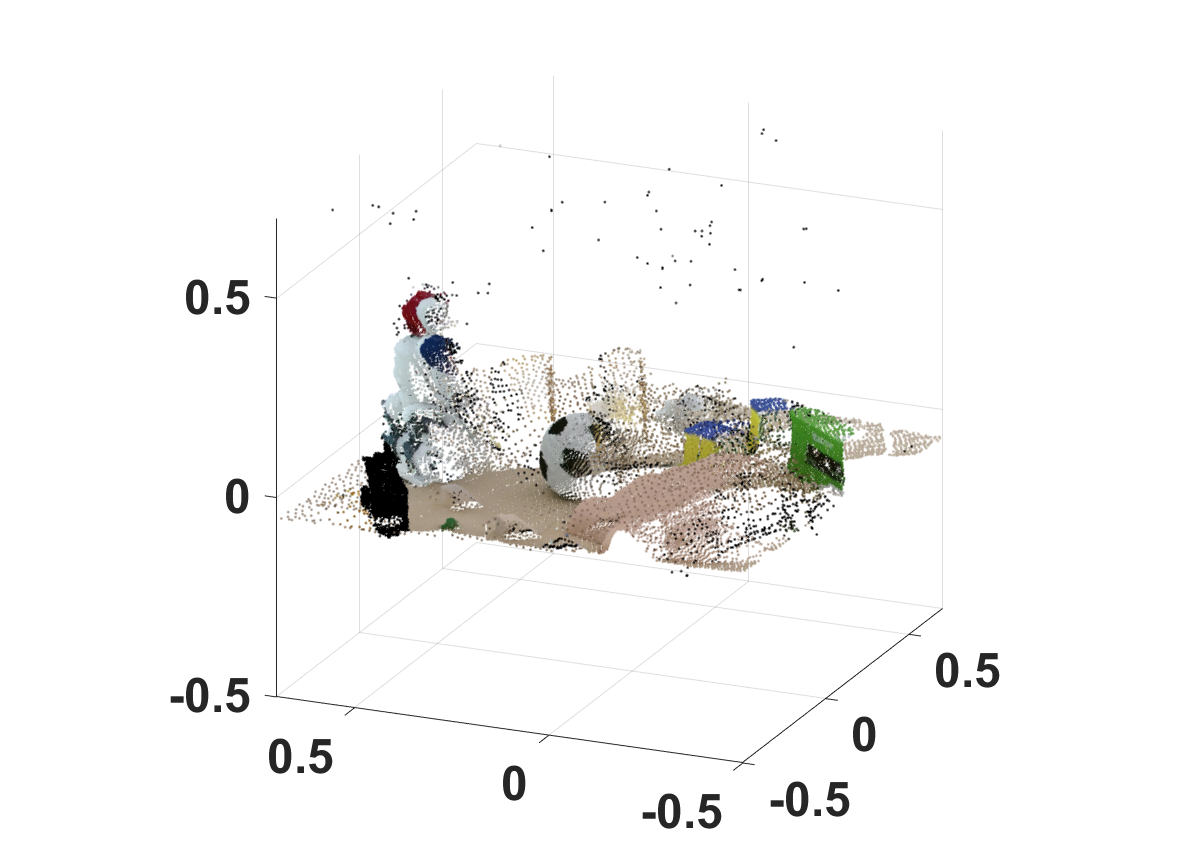}} 
	
	\caption{In this work, DUGMA incorporates the 3D uncertainty distribution of each 3D point from a sensor into a dynamic Gaussian mixture alignment system. The figure shows a critical example of aligning two  noisy and partial 3D scans with many outliers and different densities from two real low-resolution scanners using our algorithm.}
	\label{fig:Kinect}
\end{figure}

With recent improvements in depth sensing devices and algorithms, 3D point clouds are more accessible to researchers. However, using an expensive high-precision 3D scanner to get accurate and large-scale 3D point clouds is still not popular. Accurate alignment of noisy and partial point clouds with many outliers from cheap low-resolution sensors is still a core technique in various fields, such as computer vision, robotics, virtual and augmented reality.

Finding the accurate transformation between two noisy rigid point clouds is generally hard: true point-to-point correspondences seldom exist, which limits the accuracy of the methods based on ICP~\cite{GO-ICP,BDICP,GICP_segal2009,ICP_besl1992,TrICP,LM-ICP,Survey_ICP}. As for the methods based on local descriptors~\cite{FDCP,3DMATCH,FPFH,SHOT}, coarse points tend to cause inaccurate local descriptors to mismatch with each other. Also, the variable density (distant areas have a lower point cloud density) will make them unstable. Aligning probabilistic models can effectively mitigate the problems above. Thus, many researchers have been exploring different kinds of probabilistic models~\cite{BB_2017_CVPR,SVGM,SMM,GMMREG,CPD,KDE} to represent the real surface structure.

However, as far as we can tell, no one has incorporated the physical 3D uncertainty distribution information for each point from a real sensor into the probabilistic model, which allows describing the real surface structure  more accurately. The challenges mainly lie in two parts: the first is how to get the real uncertainty distribution information from the real sensors. In the recent years, an increasing number of researchers have been investigating how to estimate the uncertainty of the acquired data for different sensors, such as the Kinect sensor~\cite{Kinect_covariance_nguyen2012}, the time of flight sensor~\cite{Fusion_ToF_SV_dal2015}, the structure from motion sensor~\cite{Covariance_SFM_engel2013} and the stereo vision sensor~\cite{Fusion_ToF_SV_marin2016}. These suggest using physical noise models for each point to represent their individual occurrence probability in 3D space. The second challenge is how to use physical uncertainty information for each point from each specific view in the registration process. Specifically, if we use a Gaussian function with a covariance to represent the uncertainty of one point in the 3D space, the covariance should change with each different coordinate system in each iteration. Thus, the registration process is dynamic. Moreover, the use of the covariance estimated from different viewpoints leads to position estimates that are compatible with each physical covariance. After that, we build a bridge to make two point clouds interact with each other by sharing information, which is obviously different from traditional Gaussian mixture alignment~\cite{GMMREG,GOGMA}. The GMM of the fixed point cloud in their methods is invariant and can't share their state with the GMM of the moving point cloud, reducing the registration accuracy and also making the usage of physical uncertainty models unavailable.                 

In this paper, we propose a simple architecture combining error estimation from sample covariances and dual dynamic global probability alignment using the convolution of uncertainty-based GMMs from point clouds in the whole 3D space. Firstly, we propose an effective way to describe the change of each 3D uncertainty model in the dynamic registration process, which represents the structure of the point cloud much better. Unlike the invariant GMM (representing the fixed point cloud) in traditional Gaussian mixture alignment, we use a dynamic uncertainty-based GMM for each point set, which interact in each iteration.  To be less susceptible to local minima, we  define a more robust energy function by convolving the two dynamic GMMs over the whole 3D space rather than use time-consuming optimization methods, such as branch and bound~\cite{GO-ICP,GOGMA,BB_2017_CVPR}. The proposed dual dynamic uncertainty-based GMM's alignment can be optimized efficiently by the EM algorithm~\cite{EM_Dempster1977,Neural_bishop1995}, which experimentally shows a wider basin of convergence than other local algorithms. A new empirical approximation will be proposed to reduce the amount of calculation drastically.

The rest of this paper is organized as the following. In Section 2, key previous registration algorithms will be reviewed briefly and also recent advances of the methods for estimating the uncertainty of the acquired data from different sensors. In Section 3, the dynamic uncertainty-based Gaussian mixture alignment theory is presented. In Section 4, tens of thousands of trials have been conducted on multiple real datasets through simulation, which is more systematic testing than that done for the compared algorithms. Also, we show real application tests with most recent and advanced algorithms. Our accuracy improvement comes from the following key contributions:

{\bf Key contributions:} 

1) Incorporation of the invariant 3D uncertainty distribution information (represented by a Gaussian function with a physical covariance) into the dynamic registration; 

2) A bridge to make the two point clouds interact with each other by creating a novel point proximity weight term;

3) A more robust energy function and efficient approximation to the optimization step that greatly reduces computational complexity. 
\section{Related Work}
In 1992, Besl and McKay~\cite{ICP_besl1992} first introduced the Iterative Closest Point (ICP) algorithm to compute the rigid transformation between two point clouds by minimizing the Euclidean distance between the corresponding points. Since then, a large number of variants have been proposed and the reader could be directed to the survey \cite{Survey_ICP}. To enhance robustness to noise, Segal et al. proposed Generalized-ICP~\cite{GICP_segal2009} in 2009, which considered the probability distribution of each point but was still based on binary correpondence search. To be robust to occlusion and small partial overlap, researchers~\cite{BDICP} built bilateral correspondence using bidirectional distances.  To widen the convergence zone, GO-ICP~\cite{GO-ICP} used a branch-and-bound method to avoid getting stuck in local minima. Exact point-to-point correspondences seldom exist and the correspondence definition (two points have the minimum distance rather than are in the same place) is coarse, which makes it hard for the ICP family to achieve accurate results.

The second class of the alignment approaches is feature-based methods, which first extract and match local descriptors (e.g. FPFH~\cite{FPFH}, SHOT~\cite{SHOT}) from two point clouds and then estimate the relative pose using random sampling~\cite{FPFH}, RANSAC~\cite{RANSAC}, Hough transforms~\cite{Demisting}, etc. Recently, Zeng et al.~\cite{3DMATCH} used a siamese neural network to learn a local patch descriptor in a volumetric grid to establish the correspondences between two point clouds. Similarly, Elbaz et al.~\cite{LORAX} used a deep neural network auto-encoder to encode local 3D geometric structures instead of traditional descriptors. Lei et al.~\cite{FDCP} proposed a fast descriptor based on eigenvalues and normals computed from multiple scales to extract the local structure of the point clouds and then recovered the transformation from matches. However, they are sensitive to noisy point clouds. Additionally, the density of the point cloud influences the extraction of local descriptors and even makes them completely break down if the density is too low.  

Aligning probabilistic models which represent the structure of the point cloud can efficiently mitigate the problems above. Our method belongs to this class. One key factor to an accurate and robust registration is the data representation used. Since 2003, many approaches based on a variety of probabilistic models have been explored to represent the structure of the point cloud such as Robust Point Matching~\cite{RPM}, Kernel Correlation~\cite{KDE}, Coherent Point Drift~\cite{CPD}. In 2011, GMMREG~\cite{GMMREG} used two Gaussian mixture models with the same isotropic covariances for each point and minimized the ${L}_{2}$ distance between the two GMMs to get the transformation. The GMM which represented the fixed point cloud was regarded as invariant and thus could not receive the current registration state from the other point cloud. In 2014, Zhou et al.~\cite{SMM} proposed to use the Student's-t mixture model to represent the point cloud in the registration algorithm. In 2015, Campbell et al.~\cite{SVGM} used a Support Vector Machine to learn and construct SVGM (a GMM with non-uniform weights) to represent the point cloud. In the next year, Campbell et al. used SVGM~\cite{SVGM} and the architecture in GMMREG~\cite{GMMREG}  to  get the globally-optimal transformation using a branch and bound approach in order not to be vulnerable to local minimum. Recently, Straub et al.~\cite{BB_2017_CVPR} used a Dirichlet process Gaussian mixture (DP-GMM) and a Dirichlet process von Mises-Fisher mixture (DP-vMF-MM) to represent the geometric information of the point cloud. The mathematical probabilistic model used to represent the point cloud has become considerably complex. Nevertheless, we will use the physical 3D uncertainty distribution from real sensors to construct a simple uncertainty-based GMM to represent the structure of the point cloud, which will fit the real surface geometry better.   

The acquisition of a physical 3D uncertainty distribution for each point from a real sensor is a difficult task. With the development of different depth sensors, more effective uncertainty estimation methods for various sensors have been designed. In 2012, Nguyen et al.~\cite{Kinect_covariance_nguyen2012} used the distance and angle between the Kinect sensor and observed surface to estimate both axial and lateral noise distributions. In 2013, Engel et al.~\cite{Covariance_SFM_engel2013} used the geometric disparity error and photometric disparity error for the structure from motion sensor to estimate 3D point error. Recently, many researchers~\cite{Fusion_ToF_SV_dal2015,Fusion_ToF_SV_marin2016} have estimated the uncertainty for the ToF (Time of Flight) sensor based on the physical properties of the sensor (eg. the IR frequency). Meanwhile, ~\cite{Fusion_ToF_SV_marin2016} developed an empirical uncertainty model for the stereo vision sensor based on the relationship between the local cost and global cost.

In summary, there are previously developed methods for both robust rigid point cloud registration and modelling the 3D uncertainty distribution of the points in a 3D point cloud. This paper improves registration accuracy and robustness using an approach that combines these two themes and redesigns a dynamic Gaussian mixture alignment system using invariant 3D uncertainty information from each point cloud.
\section{Methodology}
First we introduce the change of 3D uncertainty distribution, then build a bidirectional dynamic bridge between the two point clouds, and finally introduce the framework of our math model.
Table \ref{table_symbol_Notation} lists some of the symbols and their notations.
\begin{table}[h!]	
	\caption{Symbols \& Notations} 
	\label{table_symbol_Notation}
	\begin{tabular}{ p{2cm} p{5.5 cm}  }
		\hline
		\textbf{Symbol} & \textbf{Notation} \\
		\hline
		\textbf{$\mathbf{X}, \mathbf{Y}$} & Two point clouds \\
		$D$   & Dimensionality of the point clouds\\
		$N, M$ & Number of points in \textbf{$X, Y$} point cloud\\
		\textbf{${x}_{n}$} & One point in \textbf{$X$} point cloud\\
		\textbf{${y}_{m}$} & One point in \textbf{$Y$} point cloud\\
		
		\textbf{${\Sigma}_{{x}_{n}}$}, \textbf{${\Sigma}_{{y}_{m}}$}    & Covariance for point \textbf{${x}_{n}$} and \textbf{${y}_{m}$} \\
		\textbf{$\mathbf{I}$} &   Identity matrix \\
		\textbf{$\mathbf{0}$} &   Column vector of all zeros\\
		\hline
	\end{tabular}
\end{table}
\subsection{Change of 3D Uncertainty Distribution}
We will use a Gaussian function with a covariance to represent the distribution of one point in 3D space. The specific covariance for each point represents the physical 3D uncertainty distribution for that point from a real sensor. 

(1) If a point with covariance $\Sigma_{orig}$ has been rotated by $\mathbf{R}$, then  $\Sigma$ will be $\Sigma=\mathbf{R} \Sigma_{orig} \mathbf{R'}$. 

(2) A scaling factor for the covariance of a point is proportional to the average minimum distance $\sigma$ between two point clouds to ensure that the probability of all the points in the other point cloud will not become too small when the two point clouds are far away from each other. See Algorithm~\ref{fig:algorithm}.
\subsection{Dynamic Gaussian Mixture Alignment}
The Gaussian function \(g_{\textbf{$x_n$}}(\tau)\) of the point $ x_n $ predicts the probability that  $ x_n $ appears at the position $\tau$ in its own coordinate system. Based on Gaussian weights around each point, we will define a probability-like function that not only depends on the distribution of the point (represented by isotropic or anisotropic covariance) but also whether a possible corresponding point \textbf{\(cp_{x_n}\)} in the other point cloud is nearby. We model the presence of a corresponding point by a weight function \(w_{x_n}(\tau,cp_{x_n})\) that has significant value only when a potential corresponding point \(cp_{x_n}\) from point cloud $\mathbf{Y}$ is near the position~$\tau$.  A similar definition holds for \(g_{\textbf{$y_m$}}(\tau)\), \(w_{y_m}(\tau,cp_{y_m})\). Thus either point cloud can receive and send current state information from or to the other point cloud bi-directionally to evaluate the current registration quality.

In the analysis below, we assume the $\mathbf{Y}$ point cloud has been already transformed from the initial point cloud $\bold{Y_0}$ by rotation $\mathbf{R}$ and translation $\mathbf{t}$ (which then become the domain for the optimization of the evaluation function). 
The product $g_{x_n}(\tau) g_{y_m}(\tau)$ represents $x_n$, $y_m$ appearing at the same position $\tau$ in the same coordinate system. 
Thus, it encodes the underlying prior knowledge, ie. $x_n$, $y_m$ are possible corresponding points from two point clouds.  
In other words, any two points from the fixed and moving point cloud can be a corresponding pair in our system and the likelihood depends on the probability of correspondence that  $x_n$, $y_n$ appear at the same position $\tau$, which is different from soft assignment \cite{RPM_gold1995} in essence. 

\subsection{The description of our model}
Based on the previous discussion, we design the uncertainty-based GMM as follows:
\begin{align}
&G_{\mathbf{X}}^{\mathbf{I,0}}(\tau)=\sum_{n=1}^{N}w_{x_n}(\tau,cp_{x_n}) g_{x_n}(\tau),    \\
&G_{\mathbf{Y}}^{\mathbf{R,t}}(\tau)=\sum_{m=1}^{M}w_{y_m}(\tau,cp_{y_m}) g_{y_m}(\tau),
\end{align}
where the gaussian kernels are given by
\begin{align}
&g_{x_n}(\mathbf{\tau})=\frac{1}{\sqrt{(2\pi)^D\vert {\Sigma}_{x_n}\vert}} e^{-\frac{1}{2}(\tau-x_n)^T\Sigma_{x_n}^{-1}(\tau-x_n)}  \\
&g_{y_m}(\mathbf{\tau})=\frac{1}{\sqrt{(2\pi)^D\vert {\Sigma}_{y_m}\vert}} e^{-\frac{1}{2}(\tau-y_m)^T\Sigma_{y_m}^{-1}(\tau-y_m)} \\
&w_{x_n}(\mathbf{\tau}, cp_{x_n})= e^{-\frac{1}{2}(cp_{x_n}-\tau)^T\Sigma_{x_n}^{-1}(cp_{x_n}- \tau)} \\  
&w_{y_m}(\mathbf{\tau}, cp_{y_m})= e^{-\frac{1}{2}(cp_{y_m}-\tau)^T\Sigma_{y_m}^{-1}(cp_{y_m}-\tau)}     
\end{align}
$G_{\mathbf{X}}^{\mathbf{I,0}}(\tau)$ denotes the GMM from the fixed point cloud $\mathbf{X}$ and $G_{\mathbf{Y}}^{\mathbf{R,t}}(\tau)$ represents the GMM from the moving point cloud $\mathbf{Y}$ after rotation $\mathbf{R}$ and translation $\mathbf{t}$. Thus  $y_m=\mathbf{R} y_{m0}+\mathbf{t}$, $\Sigma_{y_m }=\mathbf{R} \Sigma_{y_{m0} } \mathbf{R'}$ , $\vert\Sigma_{y_m}\vert=\vert \mathbf{R} \Sigma_{y_{m0}} \mathbf{R'}\vert=\vert \Sigma_{y_{m0}}\vert$ due to $\vert \textbf{R} \vert=1$.  $\Sigma_{y_m}^{-1}=(\mathbf{R} \Sigma_{y_{m0} } \mathbf{R'})^{-1}=\mathbf{R} \Sigma_{y_{m0}}^{-1} \mathbf{R'}$  due to $\mathbf{R} \mathbf{R'}=\mathbf{I}$.  At all times, $x_n=x_{n0}$ and $\Sigma_{x_n}=\Sigma_{x_{n0}}$.  Each point has its own covariance.

Integrating the product of the two dynamic GMMs (representing the overlapping effect of the two point clouds) over the whole 3D space, as we shall show in our experiments, makes the energy function more robust, accurate and have a wider convergence basin compared with~\cite{CPD,GMMREG,GOGMA}. 

We now formulate the optimization over rotation \textbf{R} and translation \textbf{t} as an EM-like process. First, an energy function is defined as the following
\begin{equation}
\label{eq1}
E=\int P(\mathbf{\tau})  \log \big[G_{\mathbf{X}}^{\mathbf{I,0}}(\mathbf{\tau})  G_{\mathbf{Y}}^{\mathbf{R,t}}(\mathbf{\tau})\big]\,\textrm{d}\mathbf{\tau},  
\end{equation}
where ${\mathbf\tau}$ integrates over all the domain of the point clouds; $P(\mathbf{\tau})$ is the probability that there is a point at the position~$\mathbf{\tau}$.  We design it as the sum of the probability that all the possible corresponding pairs appear at the position $\mathbf{\tau}$ in 
\begin{equation}
P(\mathbf{\tau})=\sum_{i=1}^{N}\sum_{j=1}^{M}P(\mathbf{\tau},x_i, y_j)=\sum_{i=1}^{N}\sum_{j=1}^{M}g_{x_i}(\mathbf{\tau})g_{y_j}(\mathbf{\tau}).
\end{equation}

Equation~\eqref{eq1}  can be rewritten as
\begin{equation}
\label{eq2}
E=\int P(\mathbf{\tau}) \log \big[\sum_{i=1}^{N}\sum_{j=1}^{M}F^{\mathbf{R,t}}(\mathbf{\tau},x_i,y_j)\big]\,\textrm{d}\mathbf{\tau},
\end{equation}
with a combined term
\begin{equation}
F^{\mathbf{R,t}}(\mathbf{\tau},x_i,y_j)=w_{x_i}(\mathbf{\tau}, cp_{x_i}) g_{x_i}(\mathbf{\tau}) w_{y_j}(\mathbf{\tau}, cp_{y_j})  g_{y_j}(\mathbf{\tau}). 
\end{equation}
By the definitions above, the weight term $w_{x_i}(\mathbf{\tau}, cp_{x_i})$ is nearly zero when point ${x_i}$ is far from any point in \{$y_j$\}. 
This allows us to avoid having to compute correspondences by using all $y_j$ in place of $cp_{x_i}$ (and similarly for $cp_{y_j}$) and simplify $F^{\mathbf{R,t}}$ with 
\begin{equation}
\tilde{F}(\mathbf{\tau},x_i,y_j)=w_{x_i}(\mathbf{\tau}, y_j) g_{x_i}(\mathbf{\tau}) w_{y_j}(\mathbf{\tau}, x_i)  g_{y_j}(\mathbf{\tau}). 
\end{equation}
We maximize Equation~\eqref{eq2} to get the estimated rotation and translation by minimizing its negative. We adopt the EM algorithm \cite{EM_Dempster1977,Neural_bishop1995} to solve for $\mathbf{R}$, $\mathbf{t}$. Its main idea is: guess the values of the parameters firstly in the last iteration (denoted by `old') and calculate the posteriori probability $P^{old}({x_i}, {y_j} \vert \mathbf{\tau})$ using Bayes' theorem then, which corresponds to the expectation stage. After that, minimize the expectation of the negative log-likelihood function $\mathcal{L}$ to find the parameters in the current iteration (denoted by `new'), which corresponds to the maximization stage. Thus, we get
\begin{eqnarray}
\label{eq3}
\mathcal{L} = 
-\int \sum_{i=1}^{N}\sum_{j=1}^{M} B(\mathbf{\tau},x_i, y_j) \log (\tilde{F}^{new}(\mathbf{\tau},x_i,y_j)) \, \textrm{d}\mathbf{\tau}, \\
B(\mathbf{\tau},x_i, y_j) = P(\mathbf{\tau}) \, P^{old}(x_i, y_j \mid \mathbf{\tau}).
\end{eqnarray} 

Neglecting the constant term and using $P(\mathbf{\tau}) \approx P^{old}(\mathbf{\tau}) $,  we simplify the target function to
\begin{equation}
\label{eq4}
\mathcal{L} = \sum_{i=1}^{N}\sum_{j=1}^{M} \int  P^{old}( \mathbf{\tau},x_i, y_j) \, Mah^{new}(\mathbf{\tau},x_i,y_j)\,\textrm{d}\mathbf{\tau},
\end{equation} 
where a term similar to Mahalanobis distance is obtained:
\begin{align}
Mah^{new}(\mathbf{\tau},x_i,y_j)&=\frac{1}{2}(\mathbf{\tau}-x_i)^T(\Sigma_{x_i}^{-1}+\Sigma_{y_j}^{-1})(\mathbf{\tau}-x_i) \nonumber  \\
&+\frac{1}{2}(\mathbf{\tau}-y_j)^T(\Sigma_{x_i}^{-1}+\Sigma_{y_j}^{-1})(\mathbf{\tau}-y_j).     
\end{align}

As we will justify below, there is no real benefit to integrate the whole 3D space, because the values of the Gaussian functions are only significant near the data points themselves. In fact, because most values are quite low, we approximate the integral by a sum at only the data points to speed up the algorithm drastically (unlike \cite{GOGMA}). Thus we need evaluate only each term $P^{old} (\tau, x_i,y_j) Mah^{new}(\mathbf{\tau},x_i,y_j)$ at the positions of $x_i$ and $y_j$, which will reduce the time complexity greatly to only $\mathcal{O}(MN)$. 
Applying this simplification, the approximated energy function becomes
\begin{equation}
\label{eq5}
\tilde{\mathcal{L}} = \sum_{i=1}^{N}\sum_{j=1}^{M} \sum_{{\tau \in \{x_i,y_j\}}}  
P^{old}( \mathbf{\tau},{x_i}, {y_j}) \, Mah^{new}(\mathbf{\tau},{x_i},{y_j}).
\end{equation} 
Expanding the last sum and uniting like terms we get
\begin{equation}
\label{eq6}
\tilde{\mathcal{L}} = \sum_{i=1}^{N}\sum_{j=1}^{M} C^{old}_{i,j} \cdot ({y_j}-{x_i})^T(\Sigma_{{x_i}}^{-1}+\Sigma_{{y_j}}^{-1})({y_j}-{x_i}),
\end{equation} 
where
\begin{align} \label{Cold}
C^{old}_{i,j}= &(2\pi)^{-D} |{\Sigma}_{{x_i}}|^{-\frac{1}{2}}|{\Sigma}_{{y_j}}|^{-\frac{1}{2}} \\
&\big(e^{-\frac{1}{2}({y_j}-{x_i})^T\Sigma_{{x_i}}^{-1}({y_j}-{x_i})}+e^{-\frac{1}{2}({x_i}-{y_j})^T \Sigma_{{y_j}}^{-1}({x_i}-{y_j})}\big) \nonumber.
\end{align}
The ${x_i}$, ${y_j}$, $ {\Sigma}_{{x_i}}$ and $ {\Sigma}_{{y_j}}$ in $C^{old}_{i,j}$ are from the previous iteration. 
We then minimize Equation~\eqref{eq6} over the rotation $\mathbf{R}$ and translation $\mathbf{t}$ domain, using interior point optimization~\cite{Interior_point_wright1997} as summarized in Algorithm~\ref{fig:algorithm}. 


\begin{algorithm}
	\caption{DUGMA Point Cloud Registration.}\label{fig:algorithm}
	\textbf{Input:} Two point clouds $\mathbf{X},\mathbf{Y}$ and their covariances $\Sigma_x, \Sigma_y$, 
	initial transformation $\mathbf{R} = \mathbf{I}$, $\mathbf{t} = \mathbf{0}$.
	\begin{algorithmic}[1]
		\State \textbf{EM-like optimization}, repeat until convergence:
		\Procedure{E-step}{} \Comment\textit{Update $\mathbf{Y}$, $\sigma$, $\Sigma_y$, $C^{old}_{i,j}$}
		\State $\mathbf{Y} \gets \mathbf{R} \mathbf{Y} + \mathbf{t} $
		\State $\sigma \gets \frac{1}{M} \sum_{j=1}^M d_{min}(y_j,\mathbf{X}) $ \Comment\textit{Minimum distance}
		\State $\Sigma_y \gets \sigma \;  \mathbf{R} \Sigma_y \mathbf{R'} $
		\State $C^{old}_{i,j} \gets \text{compute Eq.~\eqref{Cold}}$
		\EndProcedure
		\Procedure{M-step}{} \Comment\textit{Solve for $\mathbf{R}$,$\mathbf{t}$}
		\State Use interior point algorithm to solve  Eq.~\eqref{eq6}:
		\State $(\mathbf{R},\mathbf{t}) \gets \arg\min_{\mathbf{R,t}} \tilde{\mathcal{L}}$
		\EndProcedure
	\end{algorithmic}
\end{algorithm}

\section{Experiments}
We implemented our algorithm using Matlab and Cuda C++. We ran all the algorithms on a laptop with Intel Core i7-7820HK processor (quad-core, 8MB cache, up to 4.4GHZ) and NVidia Geforce GTX 1080 with 8GB GDDR5X. To test the accuracy and robustness of our algorithm, our proposed method is compared with relevant recent algorithms from the top journals and conferences: CPD~\cite{CPD},  GMMREG~\cite{GMMREG}, BDICP~\cite{BDICP}, GOICP~\cite{GO-ICP}, GOGMA~\cite{GOGMA}, 3DMATCH~\cite{3DMATCH}\footnote {We only compared ours with 3DMATCH in the Kinect data application with their pre-trained weights. }, FDCP~\cite{FDCP}\footnote{FDCP has compared their results with \cite{BB_2017_CVPR} so we neglected \cite{BB_2017_CVPR}. }.  All the code is directly from the authors. We did not compare ours with \cite{LORAX} because we could not get our re-implemented algorithm to work well based on their partial released code. The Stanford 3D Scanning Repository~\cite{Stanford}  and our new 3D dataset have been used to do performance comparison of the algorithms. After that, 30 real scenes with ground truth from multiple Kinect sensors in our new dataset\footnote{\url{https://github.com/Canpu999/DUGMA}} have been used to show the approach works on par or better in a real application compared with the rest. 

\subsection{Simulation}

\begin{figure*}
	\begin{center}
		\subfloat[Bunny]{\includegraphics[width = 2.8cm,height=2.8cm]{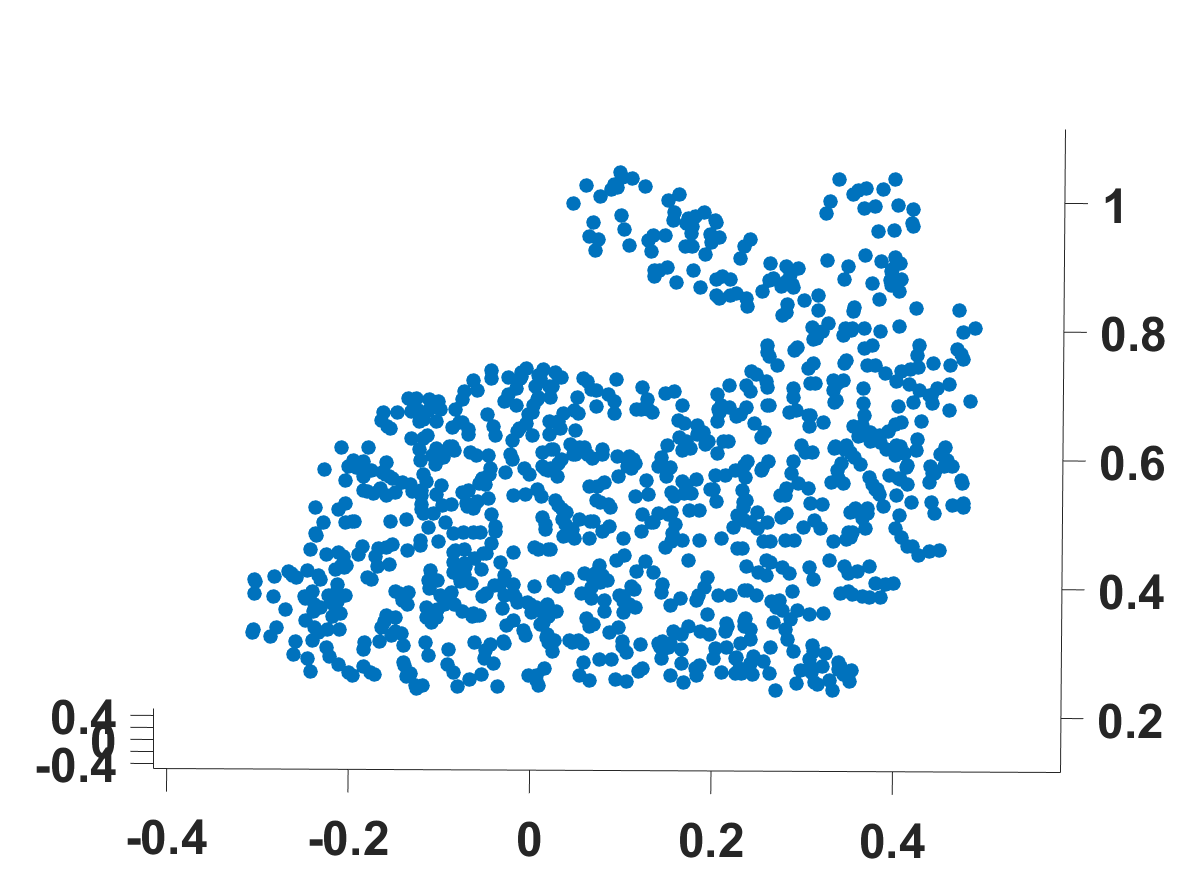}}		
		\subfloat[Armadillo]{\includegraphics[width = 2.8cm,height=2.8cm]{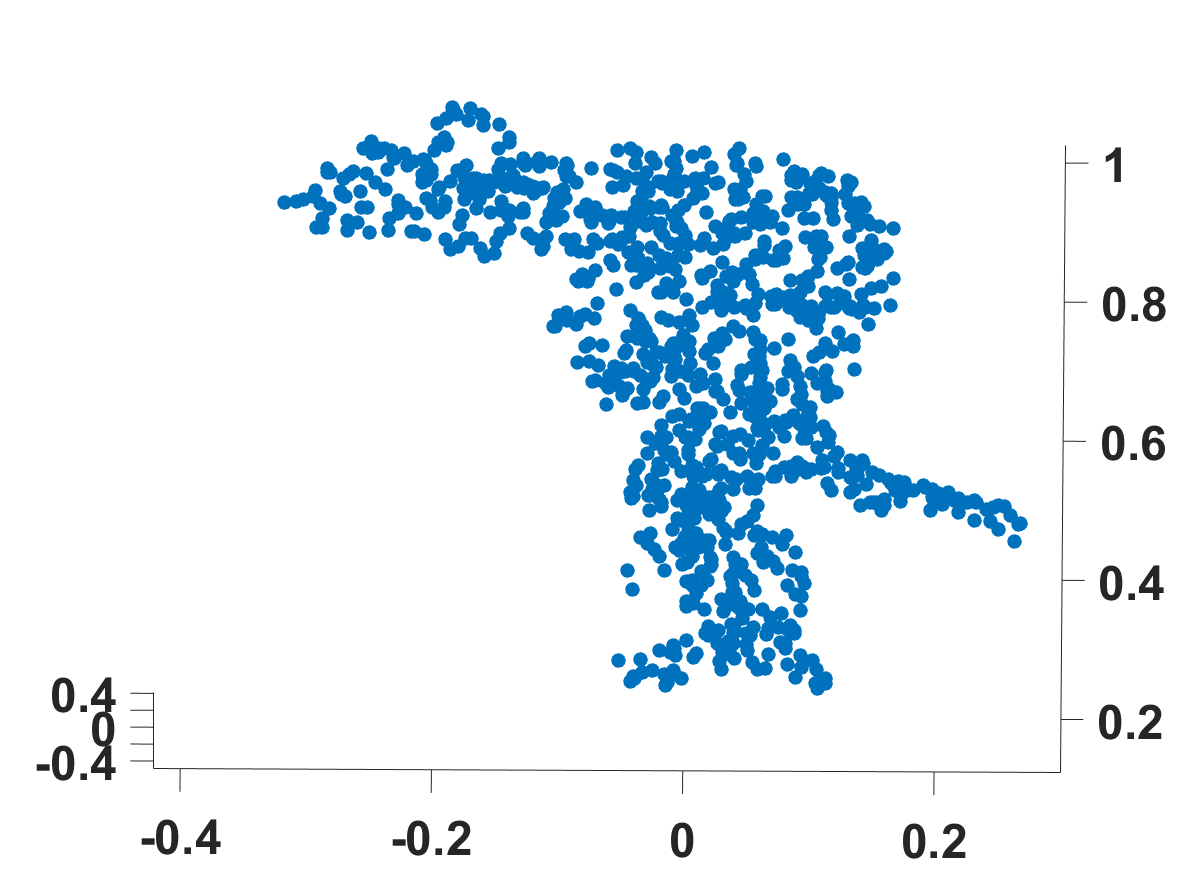}}
		\subfloat[Drill]{\includegraphics[width = 2.8cm,height=2.8cm]{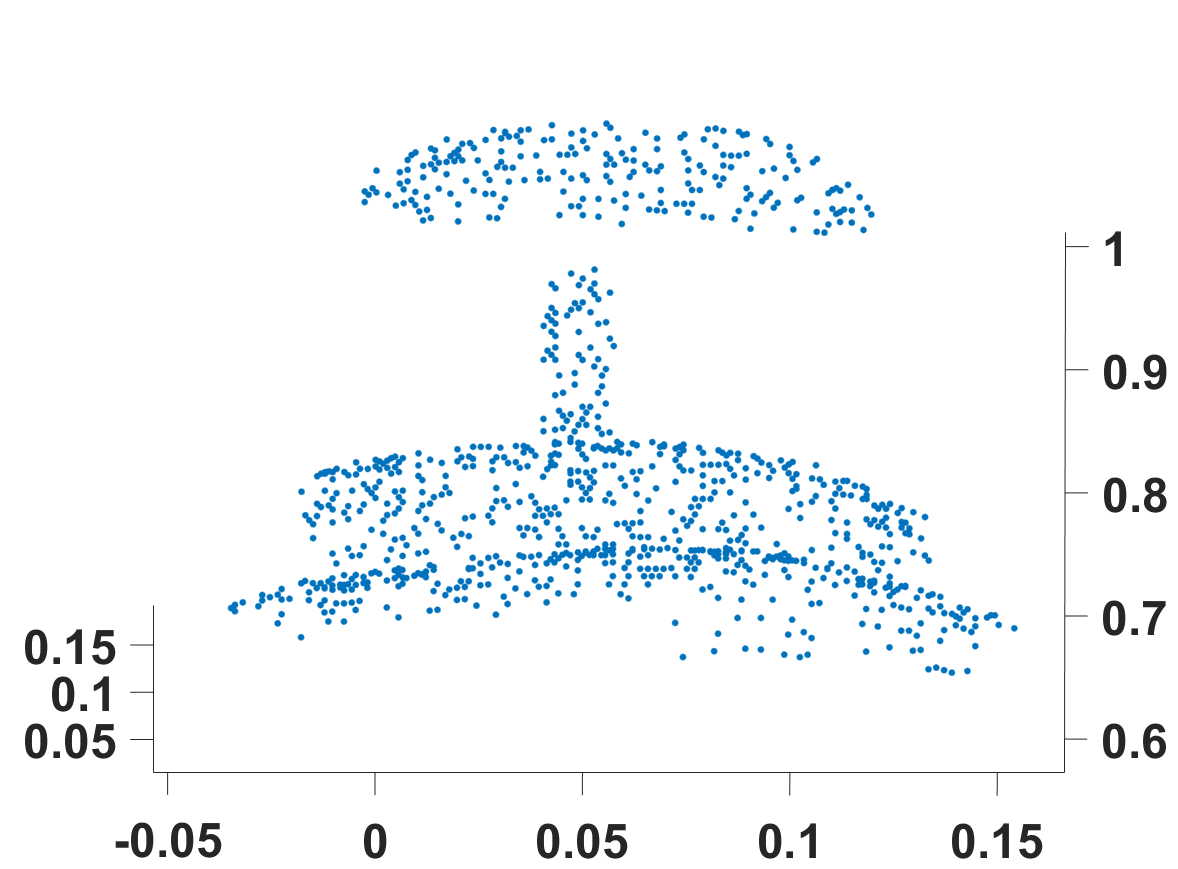}}
		\subfloat[Pole]{\includegraphics[width = 2.8cm,height=2.8cm]{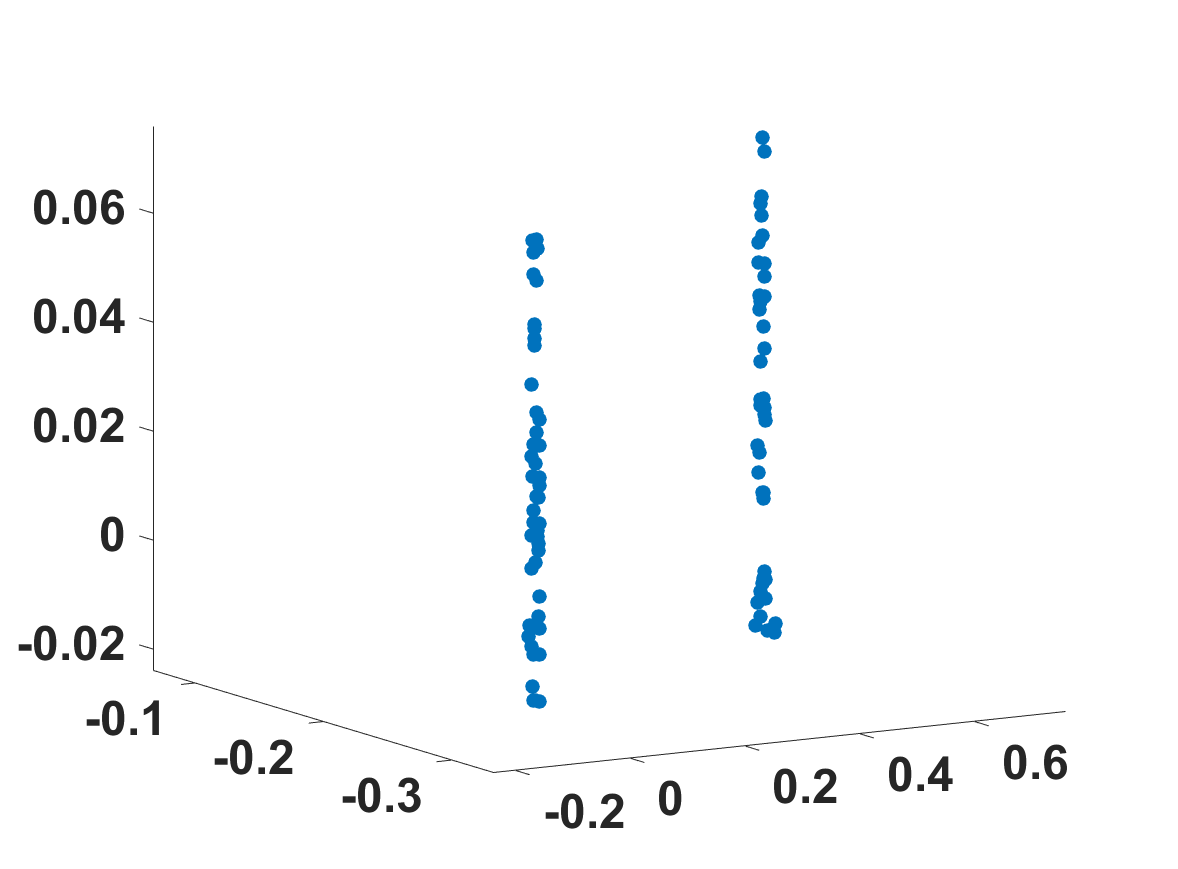}}
		\subfloat[Court]{\includegraphics[width = 2.8cm,height=2.8cm]{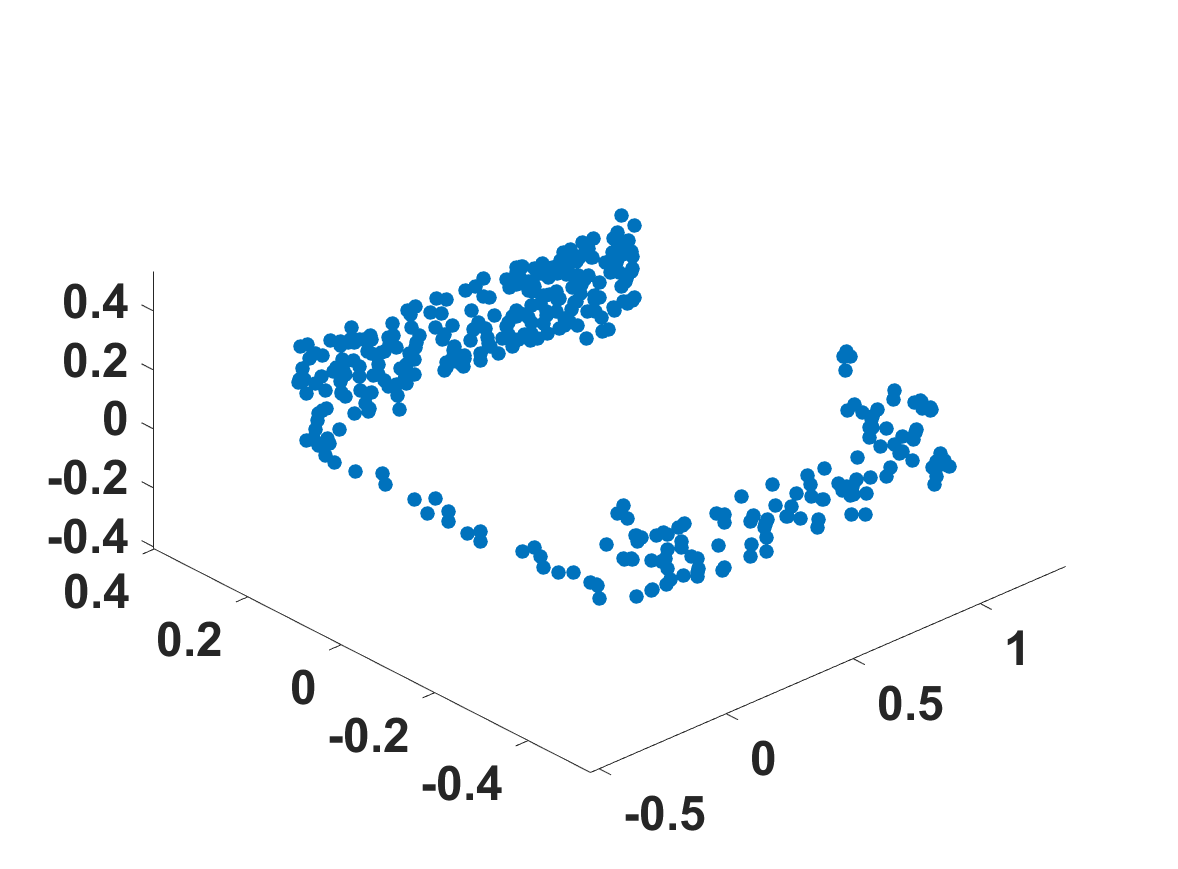}}				 	
		\subfloat[Garden]{\includegraphics[width = 2.8cm,height=2.8cm]{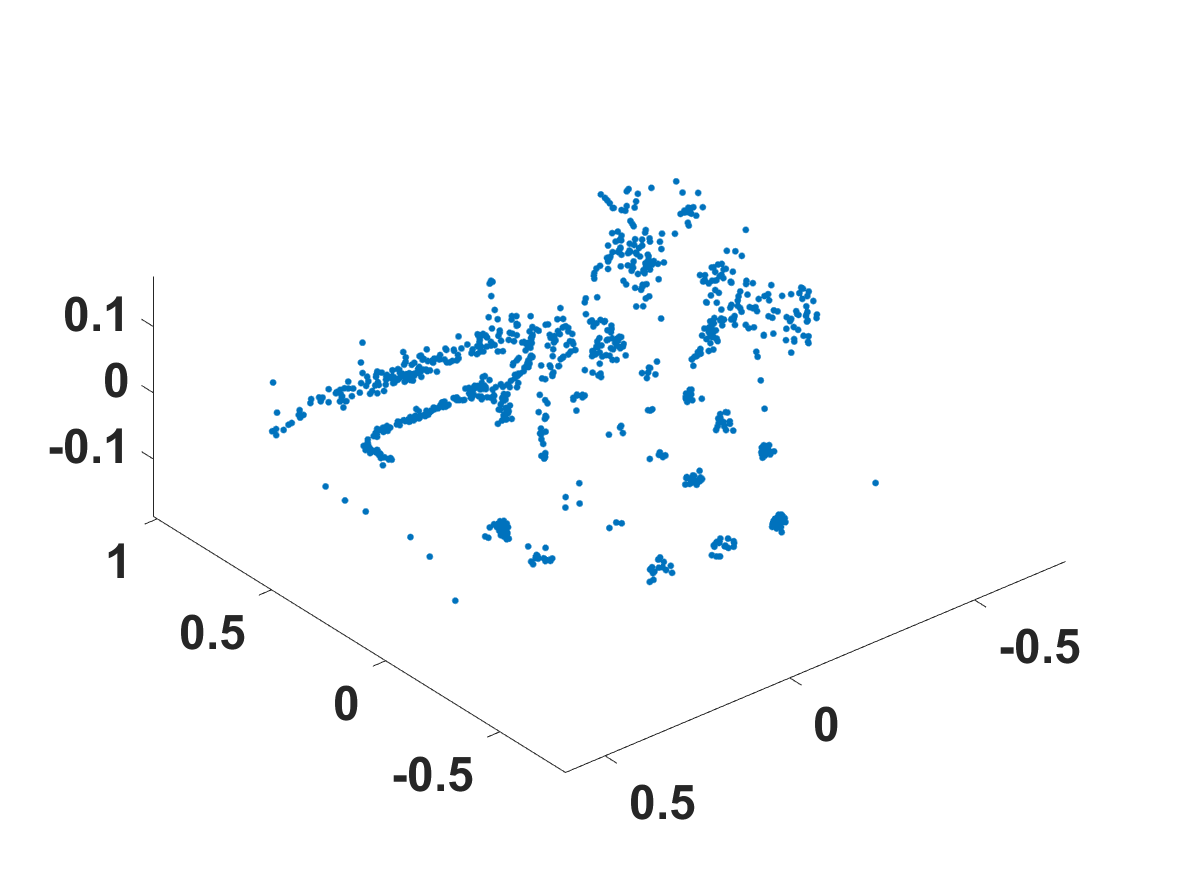}}		
	\end{center}
	\caption{six example 3D models, (a)(b)(c) from Stanford 3D Scanning Repository, (d)(e)(f) from our new dataset}
	\label{fig:3D model}
\end{figure*}
\begin{figure*}
	\begin{center}
		\subfloat[Model]{\includegraphics[width = 2.8cm,height=2.8cm]{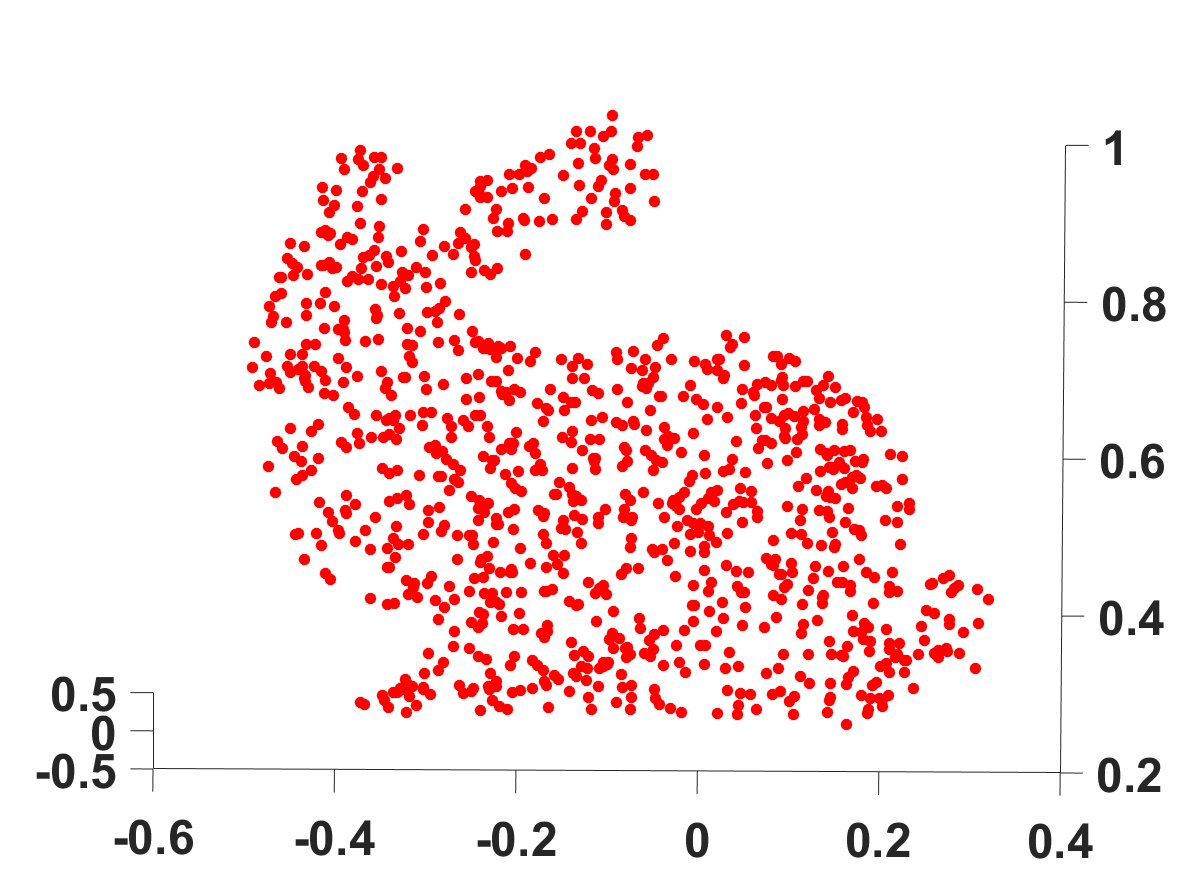}}
		\subfloat[Sampling rate 85\%]{\includegraphics[width = 2.8cm,height=2.8cm]{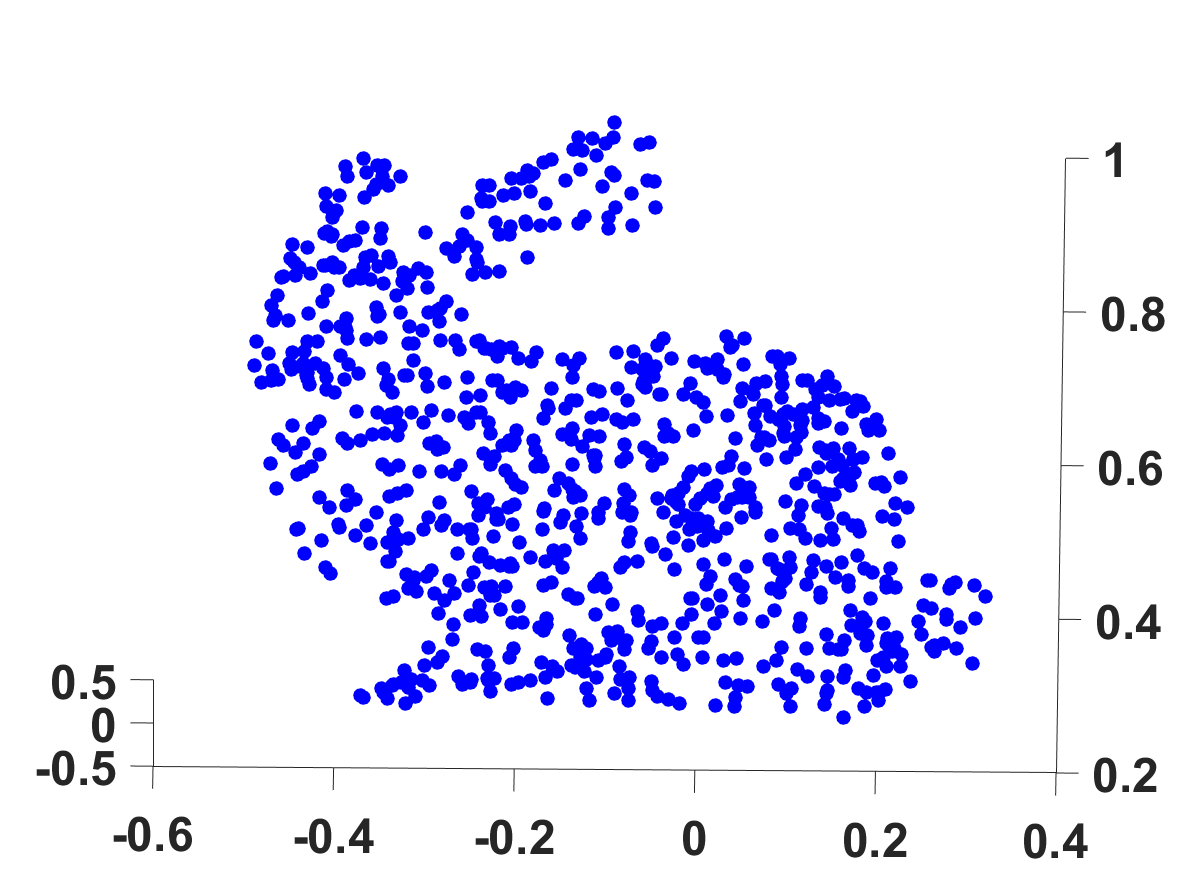}}
		\subfloat[Occlusion 10\%]{\includegraphics[width = 2.8cm,height=2.8cm]{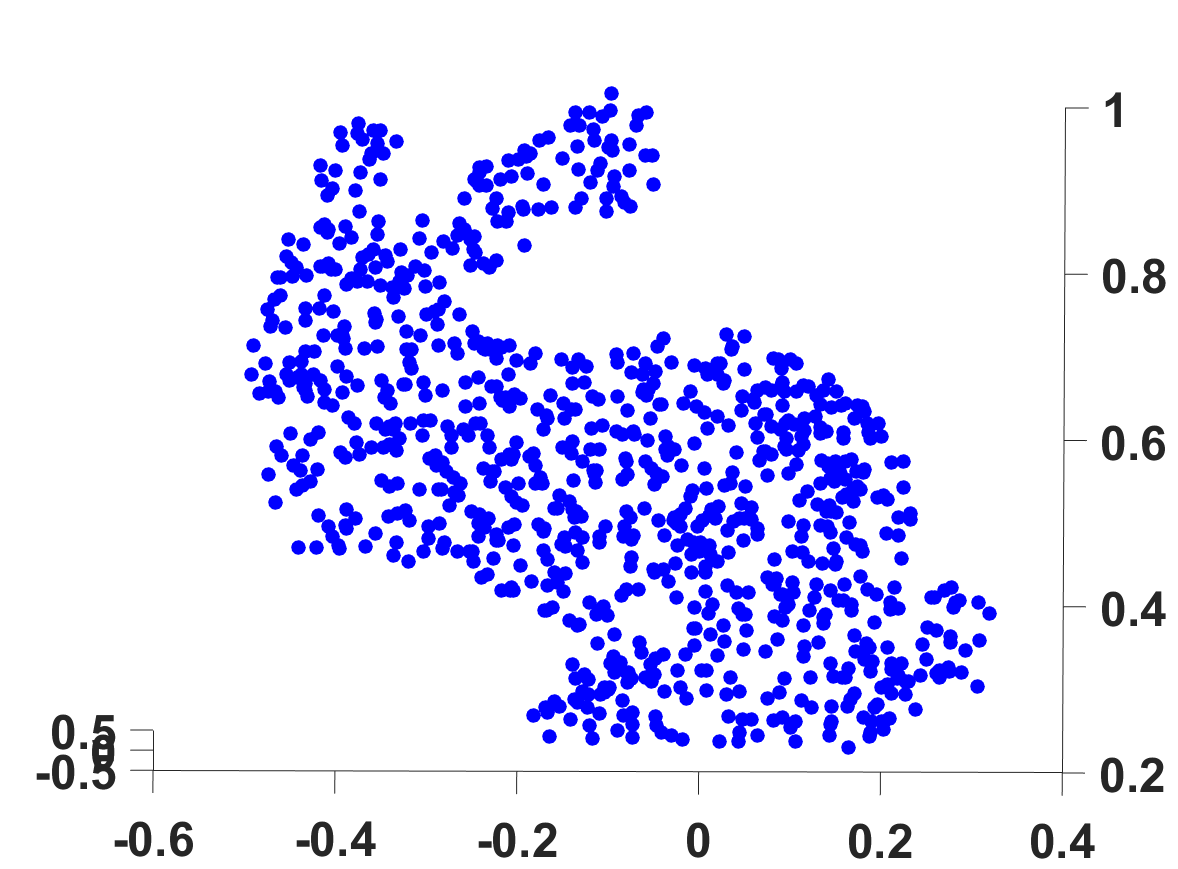}} 
		\subfloat[Outliers 200]{\includegraphics[width = 2.8cm,height=2.8cm]{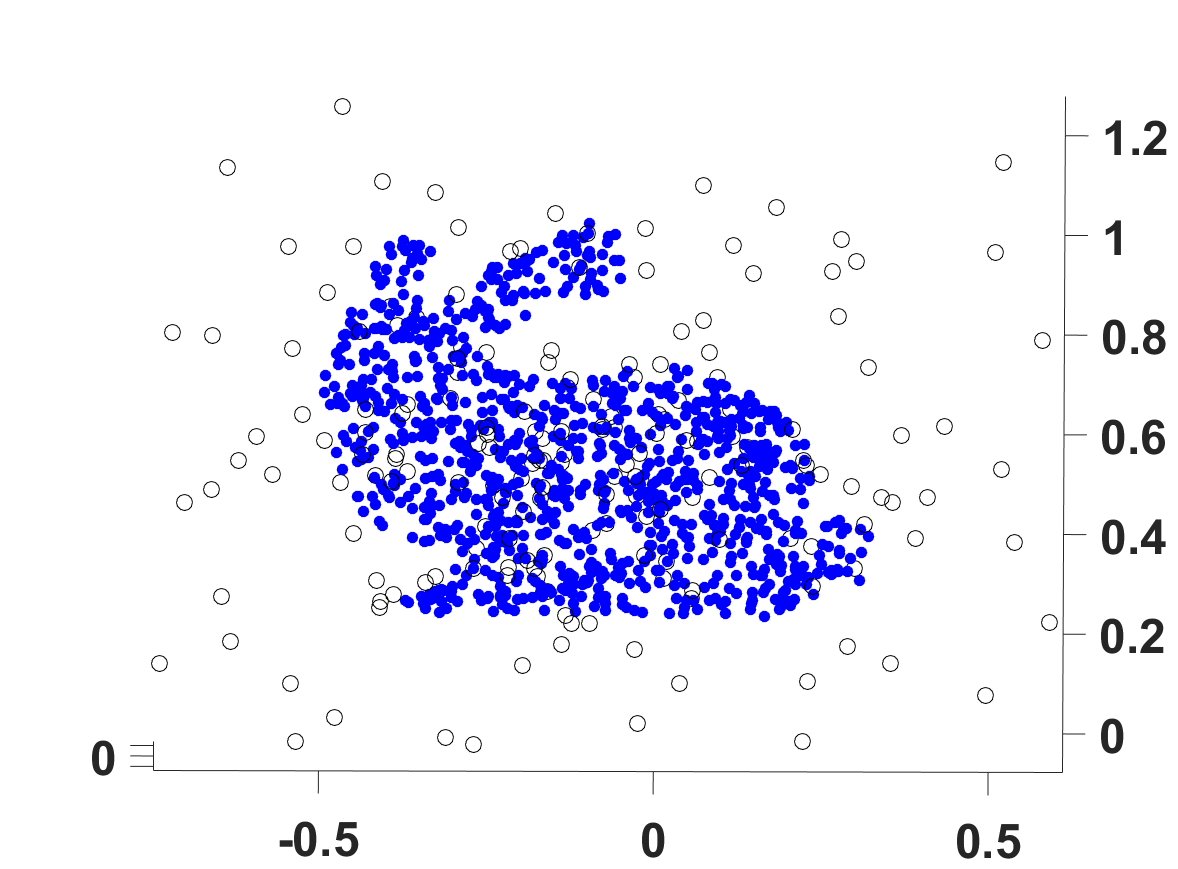}}
		\subfloat[Noise standard deviation=0.1]{\includegraphics[width = 2.8cm,height=2.8cm]{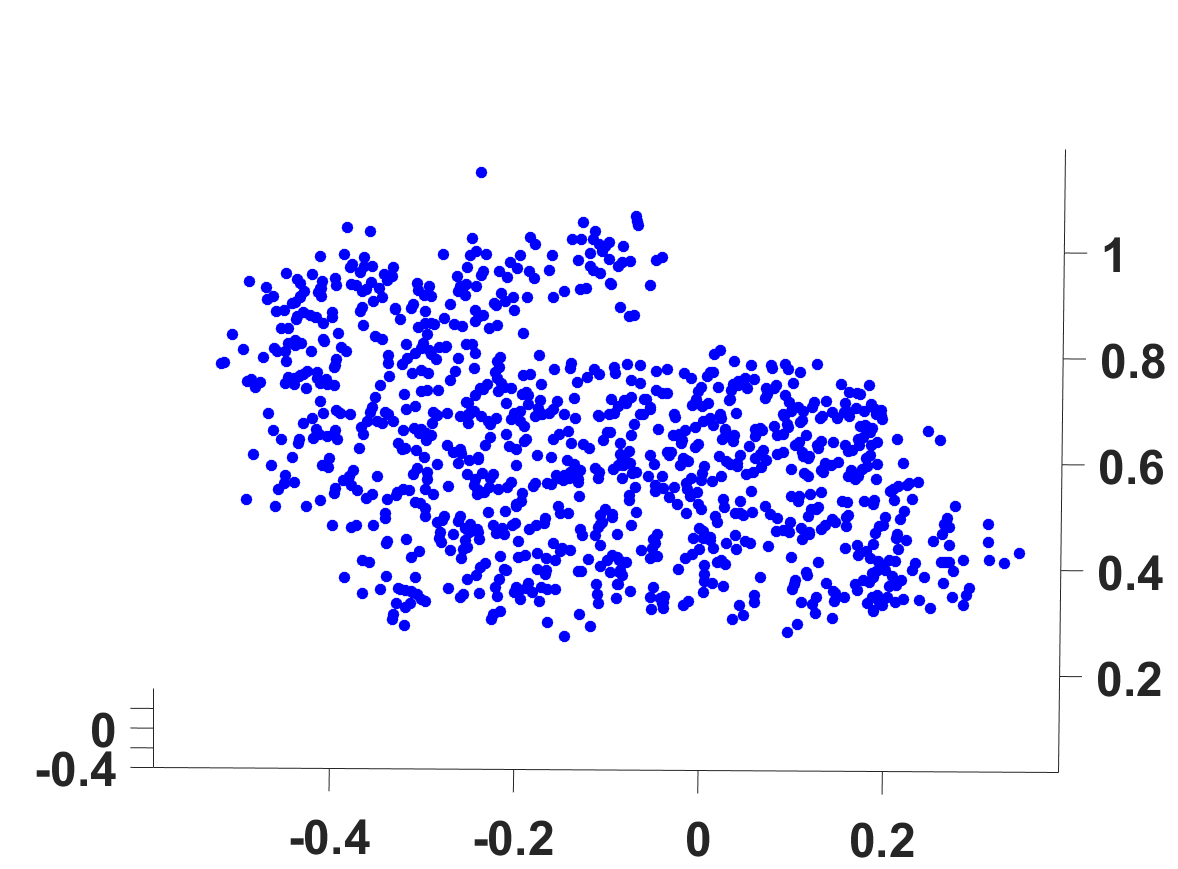}}
		\subfloat[Initial rotation angle in x,y,z=(0\textdegree,30\textdegree,30\textdegree)]{\includegraphics[width = 2.8cm,height=2.8cm]{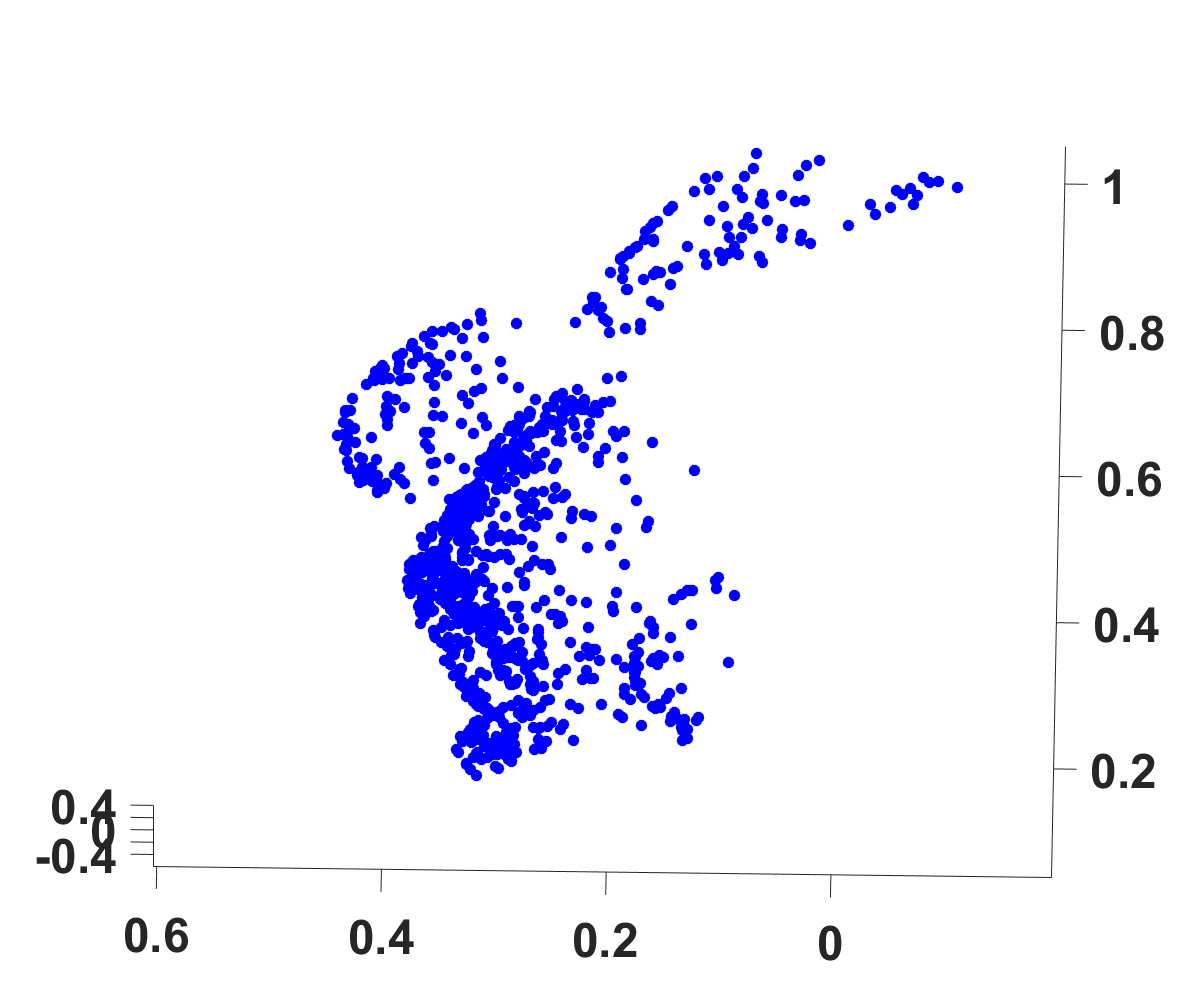}}
	\end{center}
	\caption{Different influences from various factors.}
	\label{fig:/3D_factor/}
\end{figure*}

To synthesize the two point clouds to register, we randomly choose a model from the datasets above for two point clouds firstly. Then a different random large segment of each point cloud is removed completely to simulate occlusion. After that, the two occluded models are sampled differently, which simulates the resolution of different sensors in real scenarios. Also, different anisotropic Gaussian noise with random standard deviations and zero mean has been added to each point to simulate the complex noise in real environments resulting from known and unknown factors.  The variances of all the noise
on each axis have been stored in the covariances accurately.  Next, outliers have been added into both point clouds to simulate outliers acquired by the sensors. Finally, an initial rigid transformation is applied to the moving point cloud.

The experiments are divided into four groups given the four influence factors or variables: noise, outliers, occlusion, and initial rotation. In each group of experiments, one controlled variable will be changed and the values of the other variables will be picked randomly from a default range. The experiment is conducted 3 times at each controlled value for each of 100 shapes with a random perturbation each time, see Algorithm \ref{experiment-process}. The maximum iteration value for all is 100. For FDCP, we set \texttt{gridStep}=$1.5$ and \texttt{Rho}=$0.1$ to make it robust to different densities. For GOGMA we set the scale parameter for SVM (0.5,0.5) to limit GOGMA's running time to around 100 seconds per registration. For GOICP, Mean Squared Error (MSE) convergence threshold \texttt{MSEThresh}=$0.2$. The rest of the parameters share default values in their open code.    

We use $\vert| \mathbf{t_{gt}-t_{est}}||_{F}$,  $|| \mathbf{I-R_{gt}R_{est}^{-1}}||_{F}$~\cite{Metrics_3D_Rotation_huynh2009} to estimate the quality of the registration, where $\mathbf{R_{gt},t_{gt}}$  are the ground truth and $\mathbf{R_{est},t_{est}}$ are estimated results respectively and  $|| \bullet || _{F}$ is the Frobenius norm. 

\begin{algorithm}
	\caption{Controlled and random variables process. For each method, 14700 trials have been done.}\label{experiment-process}
	\begin{algorithmic}[1]
		\For{$controlled\_variable := start \textbf{ by } step \textbf{ to } end$}
		\For{$shape := 1 \textbf{ to } 100$}
		\For {$instance := 1 \textbf{ to } 3$}
		\State Produce data with controlled and random 
		\State ~~~~variable
		\State Do registeration (different algorithms)
		\State Calculate the registration error
		\EndFor
		\EndFor
		\EndFor
	\end{algorithmic}
\end{algorithm}

From the Stanford 3D Scanning Repository (50) and our new dataset (50) we got 100 models from various views of different objects and scenes. Each was downsampled to about 1000 points with different densities. Figure \ref{fig:3D model} shows 6 example models from different scenes and objects.

We apply different effects to simulate the real factors in the real environment. Figure~\ref{fig:/3D_factor/} shows examples of the effects. In our experiment, the sampling rate is set to 90\% and 85\% for the fixed and moving point cloud, respectively. Table~\ref{3D random parameters} gives specific information about the parameters.
\begin{table}[h!]
	\centering
	\caption{Range for random and controlled factors} \label{3D random parameters}
	\begin{tabular}{|p{2.1cm}|| p{2cm}| p{2.8cm}|}
		\hline 
		\textit{Factor}  &  \textit{random range} & \textit{controlled range} \\
		\hline 
		Initial rotation  &   [-20\textdegree, 20\textdegree] & [-60\textdegree, 60\textdegree]; step=8\textdegree \\
		\hline 
		Outliers & [0, 500] that is, [0, $\approx$33\%] & [0, 2000] that is, [0, $\approx$67\%];	step=200  \\ 
		\hline 
		Noise standard deviation & [0, 0.2] $\times$ radius of point cloud & [0, 0.3] $\times$ radius of point cloud; step=0.03\\ 
		\hline 
		Occlusion & [0, 15\%] & [0, 30\%]; step=0.03\\ 
		\hline 
	\end{tabular} 	
\end{table}

\begin{figure}[h!]
	\centering
	\subfloat{\includegraphics[width = 4cm,height=4cm]{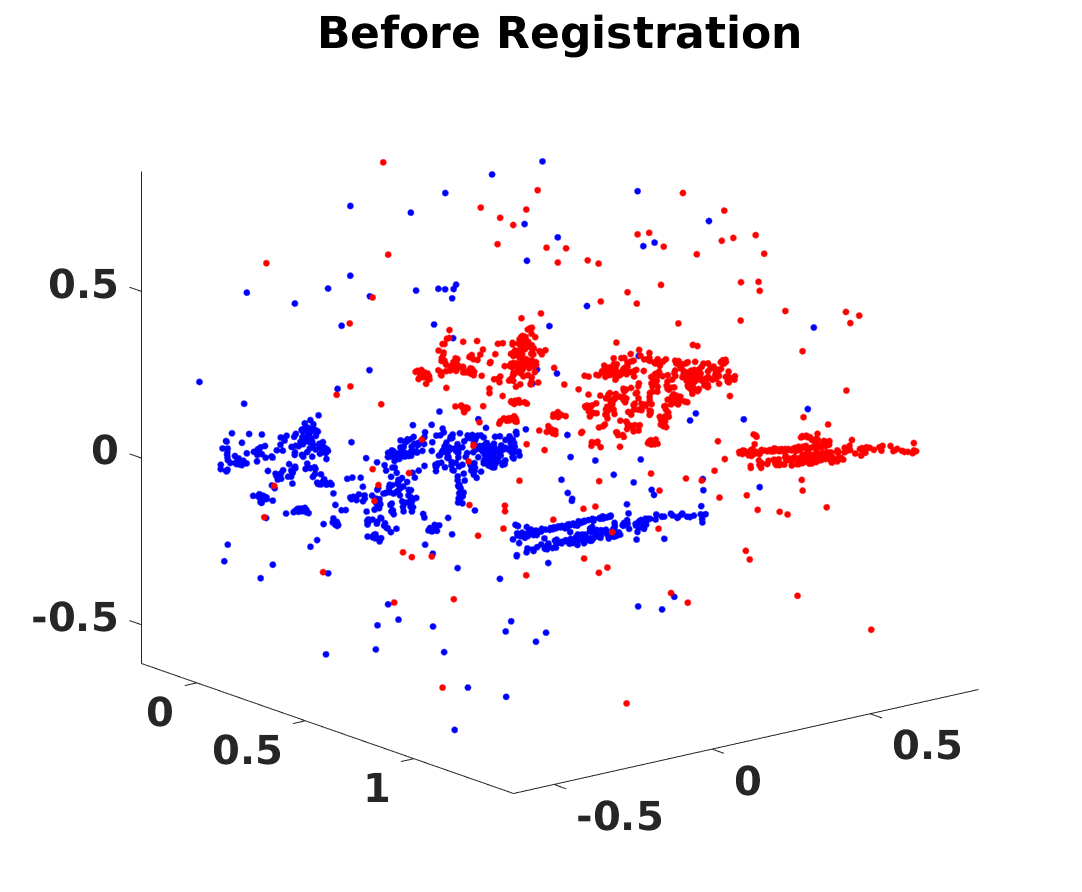}}
	\subfloat{\includegraphics[width = 4cm,height=4cm]{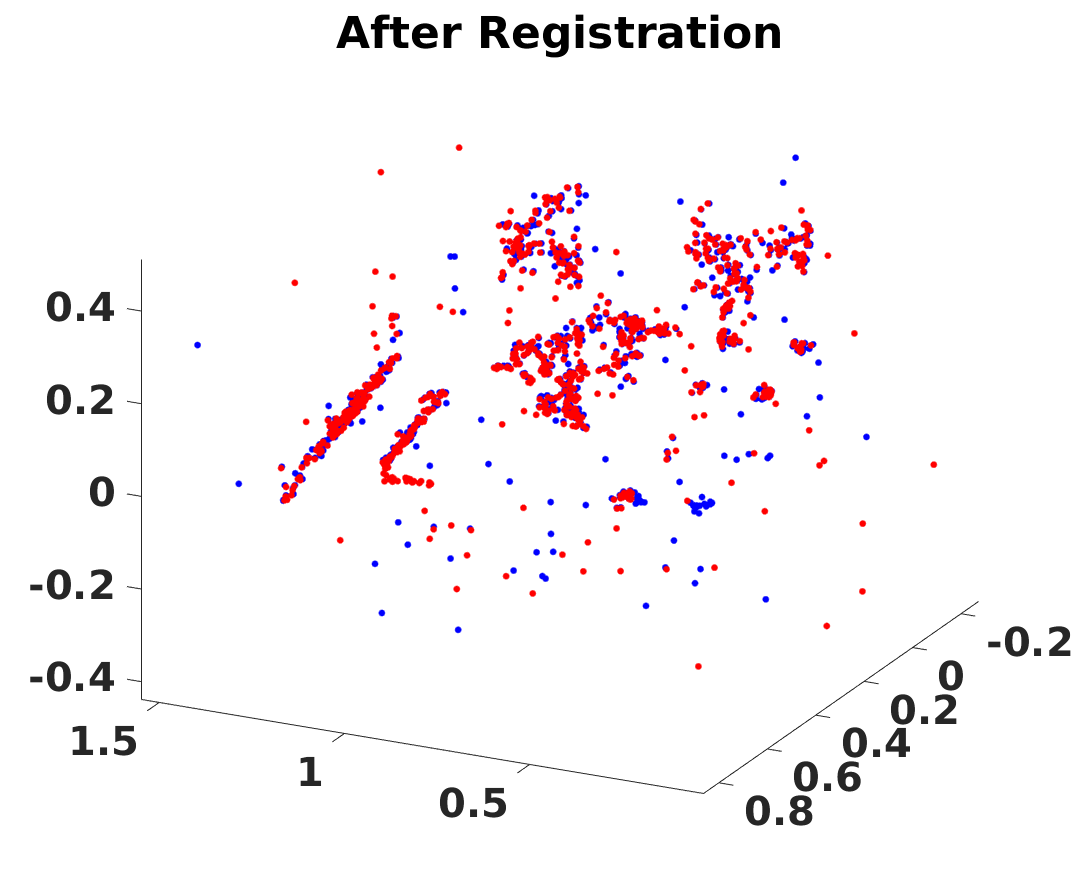}}
	
	\caption{A successful registration in a real garden.}
	\label{fig:3D registration}
\end{figure}

Figure~\ref{fig:3D registration} shows one successful registration in a real garden. After registration, we could see the hedges and trees overlap well although there is a big patch of occlusion in both two point clouds, many outliers and noise.

\begin{figure}[h!]
	\centering
	\subfloat{\includegraphics[width = 4cm,height=3cm]{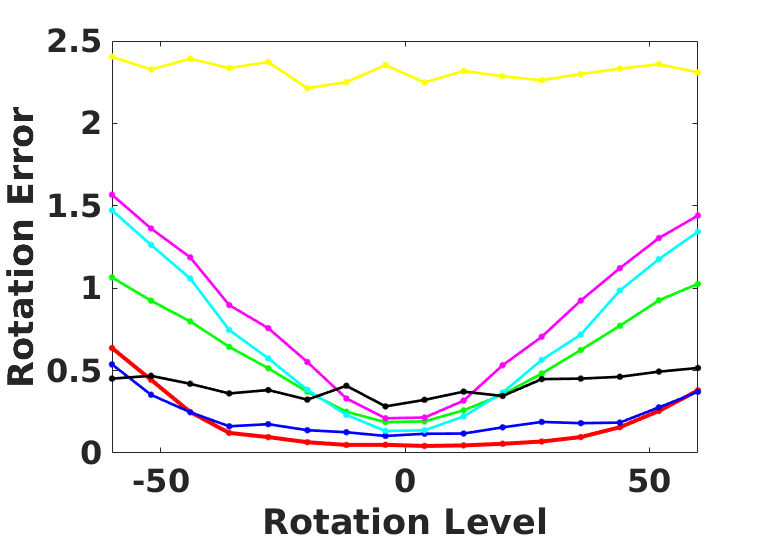}}
	\subfloat{\includegraphics[width = 4cm,height=3cm]{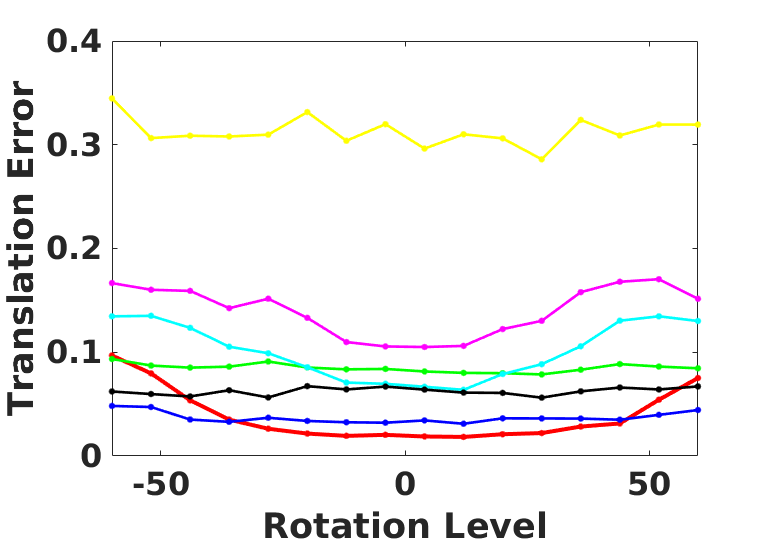}} \\
	
	\subfloat{\includegraphics[width = 4cm,height=3cm]{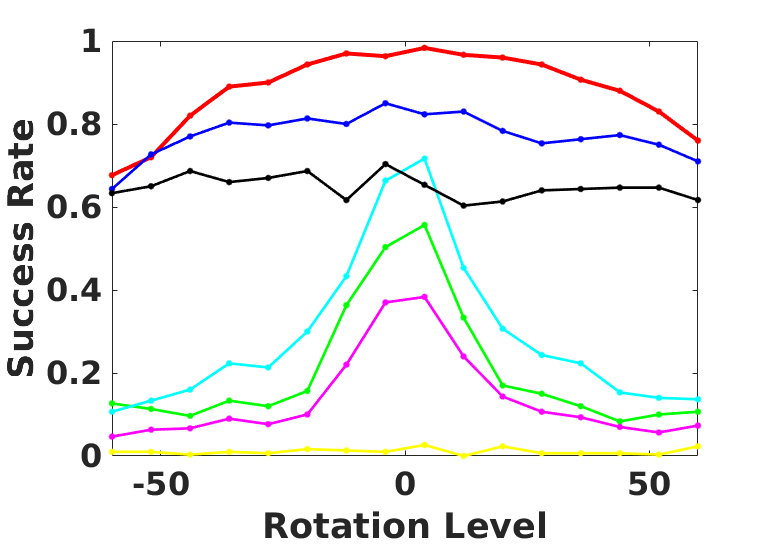}}
	\subfloat{\includegraphics[width = 4cm,height=3cm]{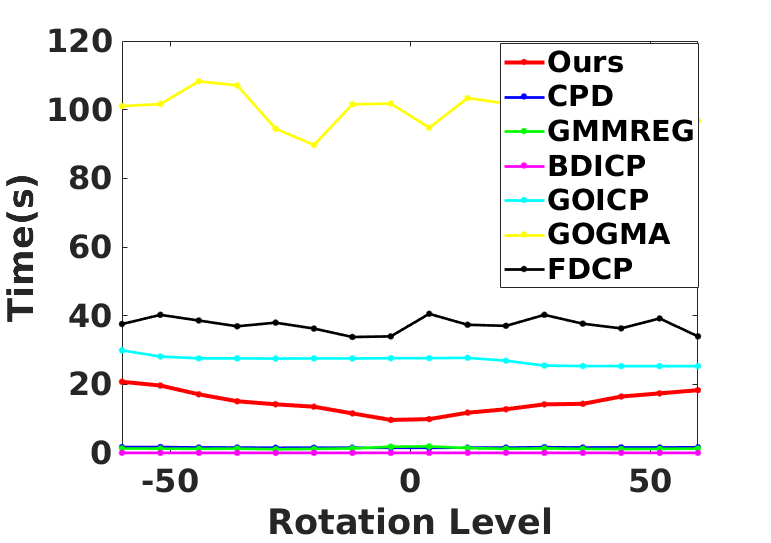}} \\
	
	\caption{Rotation Experiment}
	\label{fig:3Drotation}
\end{figure}

\begin{figure}[h!]
	\centering
	\subfloat{\includegraphics[width = 4cm,height=3cm]{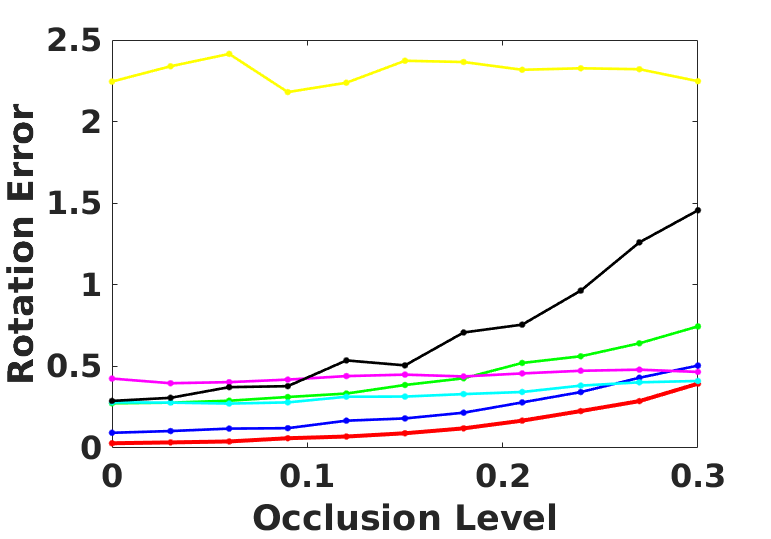}}
	\subfloat{\includegraphics[width = 4cm,height=3cm]{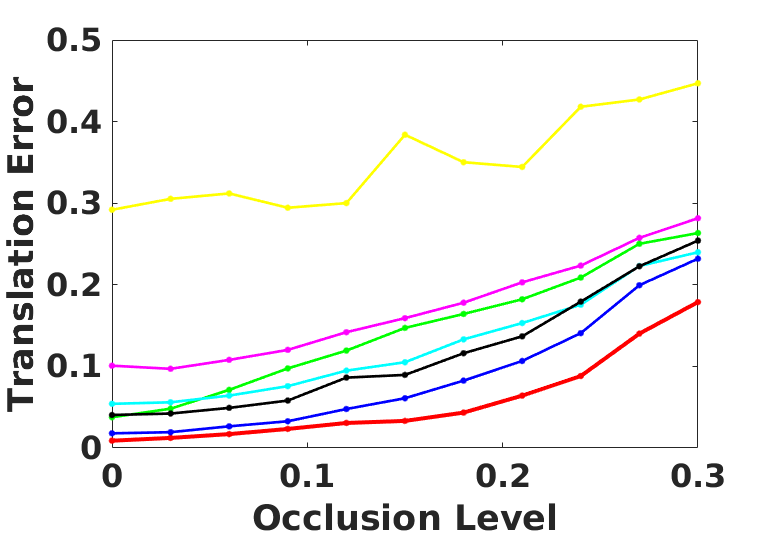}} \\
	
	\subfloat{\includegraphics[width = 4cm,height=3cm]{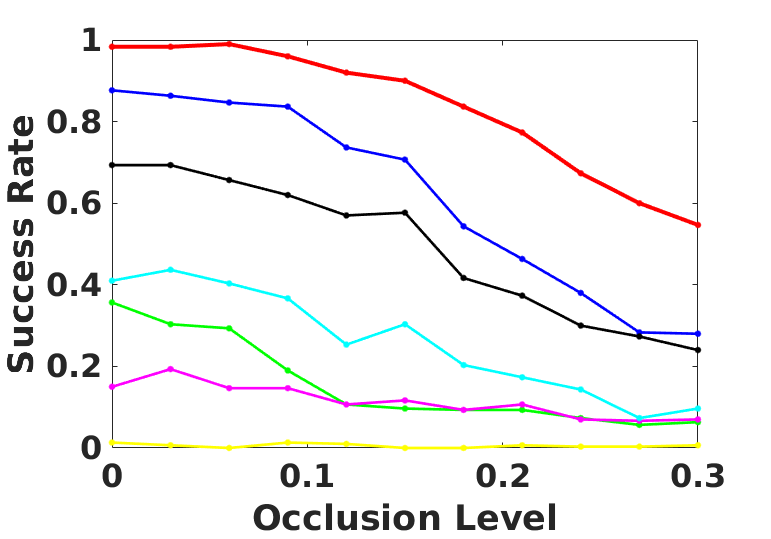}}
	\subfloat{\includegraphics[width = 4cm,height=3cm]{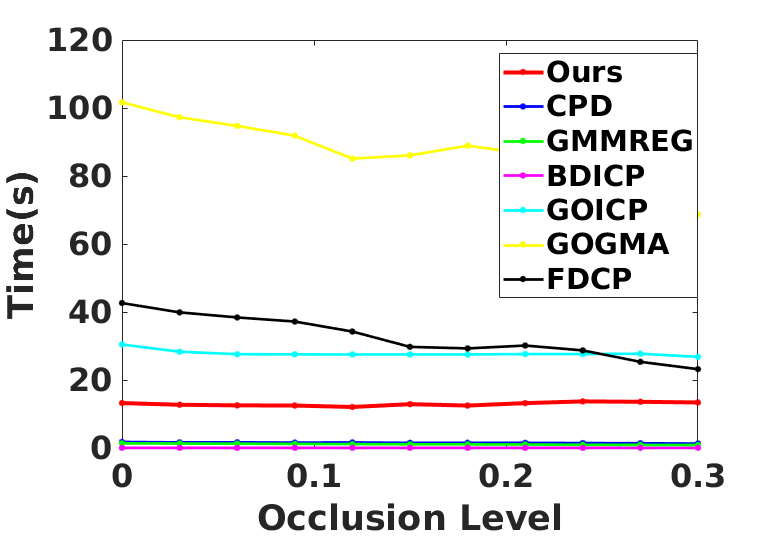}} \\
	
	\caption{Occlusion Experiment}
	\label{fig:3Docclusion}
\end{figure}

\begin{figure}[h!]
	\centering
	\subfloat{\includegraphics[width = 4cm,height=3cm]{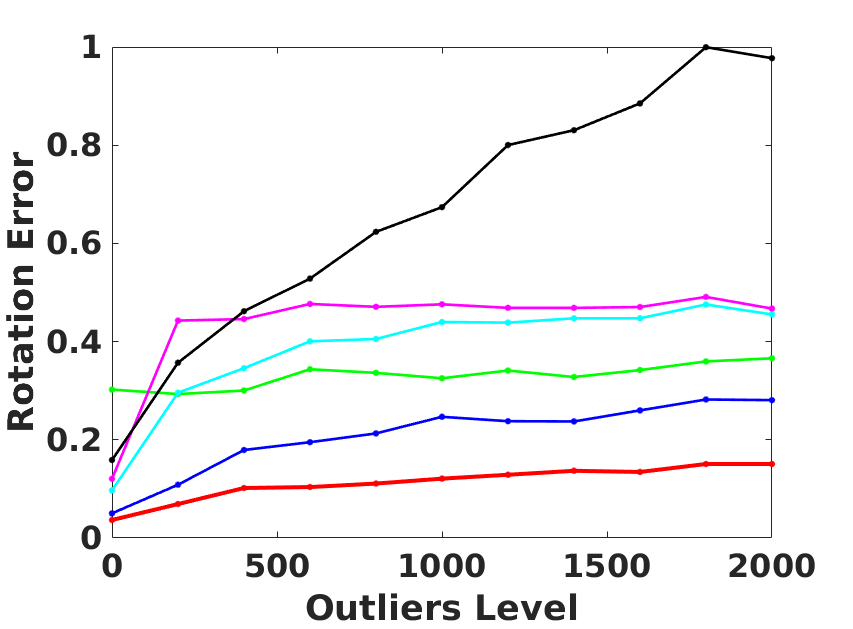}}
	\subfloat{\includegraphics[width = 4cm,height=3cm]{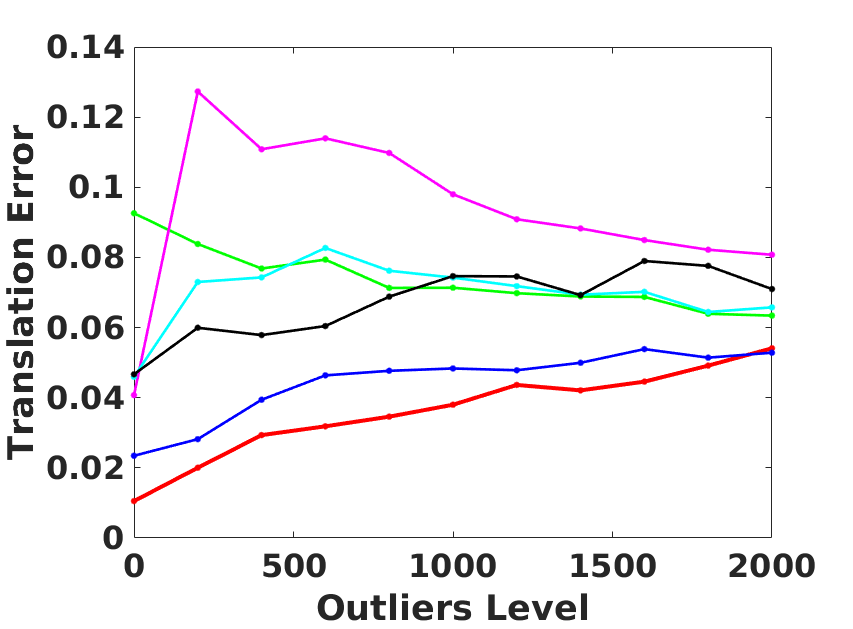}} \\
	
	\subfloat{\includegraphics[width = 4cm,height=3cm]{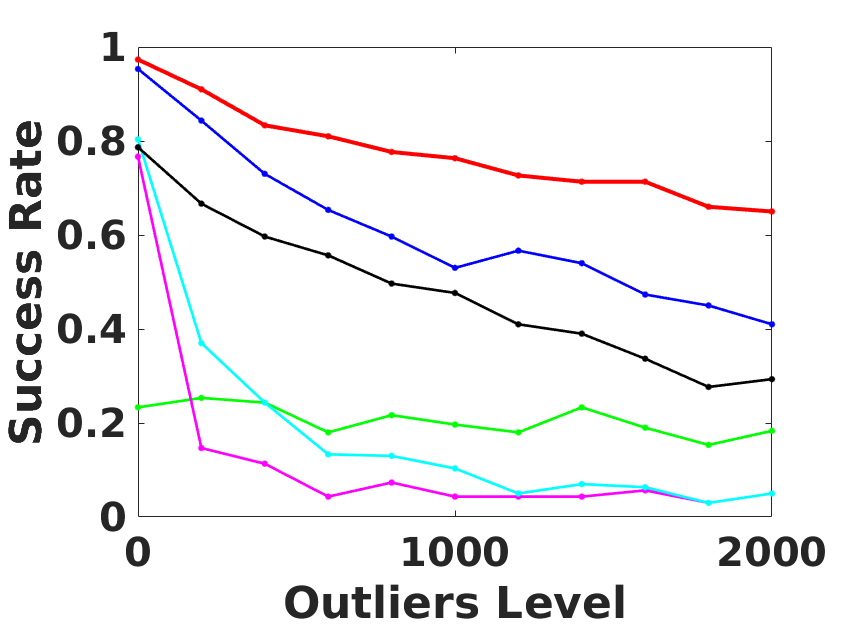}}
	\subfloat{\includegraphics[width = 4cm,height=3cm]{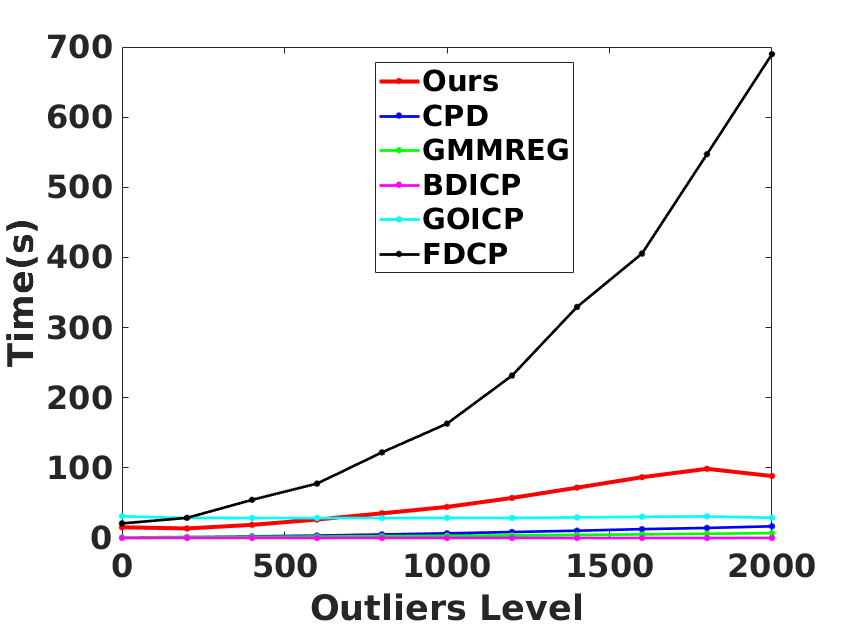}} \\
	
	\caption{Outlier Experiment}
	\label{fig:3Doutliers}
\end{figure}

\begin{figure}[h!]
	\centering
	\subfloat{\includegraphics[width = 4cm,height=3cm]{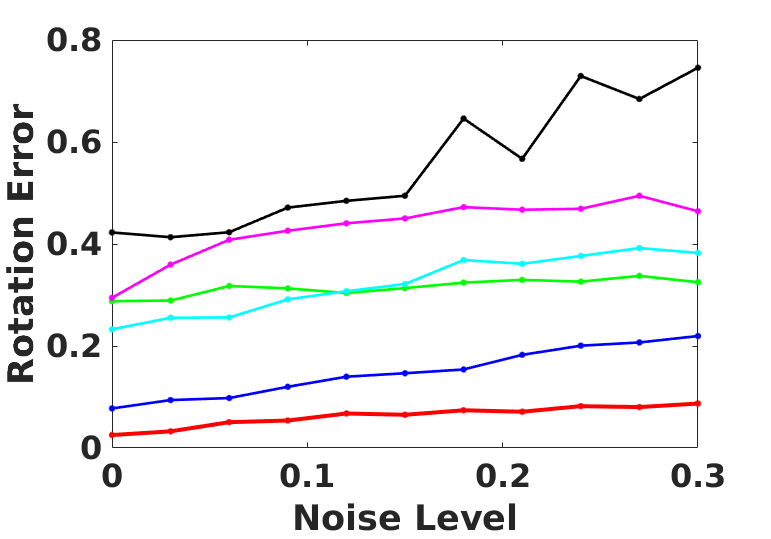}}
	\subfloat{\includegraphics[width = 4cm,height=3cm]{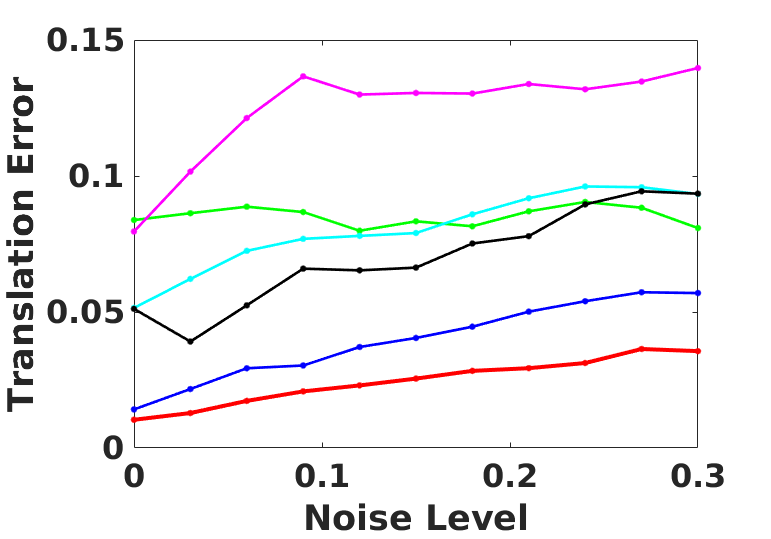}} \\
	
	\subfloat{\includegraphics[width = 4cm,height=3cm]{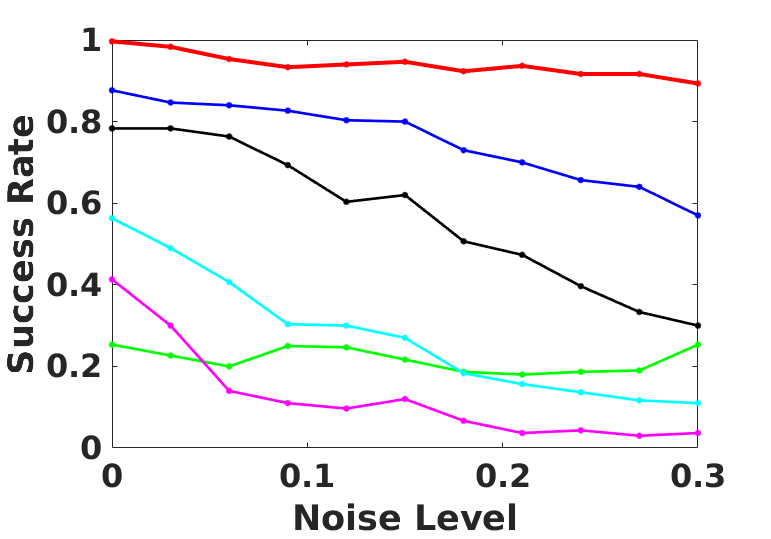}}
	\subfloat{\includegraphics[width = 4cm,height=3cm]{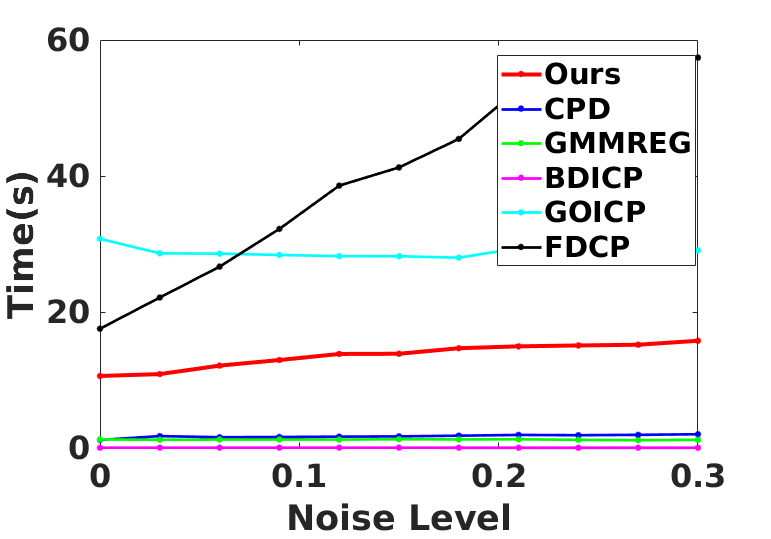}} \\
	
	\caption{Noise Experiment}
	\label{fig:3Dnoise}
\end{figure}

When the initial rotation angle value is the controlled variable, it ranges from [-60\textdegree, 60\textdegree], with an 8\textdegree~step. In the experiments, the specific rotation angle around each axis is chosen as 0 or the initial rotation angle value randomly. Figure \ref{fig:3Drotation} shows  
that beyond -40\textdegree or 40\textdegree, the proposed algorithm breaks down because the iteration count exceeds the maximum. But within [-40\textdegree, 40\textdegree], our algorithm is much more stable and accurate compared with the rest. In Figures~\ref{fig:3Drotation},\ref{fig:3Docclusion},\ref{fig:3Doutliers},\ref{fig:3Dnoise}, `Time' refers to the average running time per registration. If rotation and translation error is below 0.2 and 0.1 respectively in a trial, the trial is a success (third plot). 
When occlusion rate is the controlled variable, Figure \ref{fig:3Docclusion} shows within 25\%, the proposed algorithm performs well. 

Judging that GOGMA needs much more time (about 1000 sec for a trial) to achieve a much better performance and behaved poorly in the experiments above, we will neglect GOGMA in the remaining noise and outlier experiments but later compare the proposed algorithm with it in the small dataset registration experiment.

When outliers are the controlled variable, we use covariances generated in the same manner as the true data points. Figure \ref{fig:3Doutliers} shows the proposed algorithm has superior performance again. When the noise level is the controlled variable, Figure \ref{fig:3Dnoise} shows robust and accurate performance compared with the rest. 

\subsection{Real data from multiple Kinect sensors}
The simulation experiments above show our algorithm works well with very accurate covariance estimates. In the real case, it is hard to get very accurate covariance estimates. In this real application, we estimate an inaccurate covariance for each 3D point from a Kinect sensor to test our algorithm. We design the uncertainty of each valid 3D point acquired by the Kinect sensor based on the depth value $d$ and the angle $\alpha$ between the camera and the normal of the surface \cite{Kinect_covariance_nguyen2012}. 
\begin{equation}
\label{eq7}
U(\alpha,d)=exp[w_1 (1-\cos{\alpha} )+w_2 d]
\end{equation}

We use $w_1=1.6658$ and $w_2=0.2776$ by letting $U(\pi/3,0)=U(0,3)=2.2$. The number 2.2 is set manually and the algorithm works well if that number is in the range [1,10] (known by our experiments). Then we simply multiplied the uncertainty and the identity matrix to estimate a coarse covariance for each point. Future work will explore more accurate real covariances to represent the 3D uncertainty distribution.

We tested our algorithm using two point clouds from two Kinect sensors. The ground truth of the rotation and translation between the two Kinect sensors is known by calibration. Figure \ref{fig:Kinect} (a), (b) show the scene before and after registration using our algorithm. Figure \ref{fig:Kinect} (c) adds the colour texture information into the two registered point clouds. 

In the experiment, the two point clouds have been downsampled to $\sim$4K points or so using the grid average method. The downsampling was small from $\sim$20K to $\sim$4K (rather than 640*480 to $\sim$4K): the Kinect scan was cropped to the fixed scenes. Then we applied the same initial rotation to all the algorithms and reduced the scale parameter  for SVM (0.08, 0.08) in GOGMA and \texttt{MSEThresh}=$0.001$ in GOICP to make them get their best performance. Here we present results from 30 scenes in our new dataset. We calculated the error and running time based on only successful registrations (rotation and translation error is below 0.2 and 0.1). After all the algorithms have converged, our successful rate (most important) ranks first (96\%). The estimated mean rotation of our algorithm (0.04) is lowest, see Table \ref{Kinect time}. Otherwise, our algorithm is about 7 times faster than GOGMA whose success rate (93\%) ranks second.
\begin{table}[h!]
	\centering
	\caption{Experiment Results for the Real Application} \label{Kinect time}
	\begin{tabular}{|p{1.1cm}| p{0.7cm}|p{0.7cm}|p{0.7cm}| p{0.7cm}|p{0.7cm}|p{0.7cm}|}
		\hline
		Method & Error Mean \bf {($\mathbf{R}$)} & Error Std \bf {($\mathbf{R}$)} & Error Mean \bf {($\mathbf{t}$)} & Error Std  \bf {($\mathbf{t}$)} & Suc. rate & Time s/trial \\
		\hline 
		CPD &   0.05 & 0.03 & \bf 0.03 & \bf 0.01 & 50\% & 42.0\\
		\hline 
		Gmmreg & - & - & - & - & 0 & - \\
		\hline 
		BDICP & - & - & - & - & 0 & - \\ 
		\hline 
		GOICP &   - & - & - & - & 0 & - \\ 
		\hline  
		GOGMA &   \textbf{0.04} &  0.03 & \bf 0.03 & 0.02 & 93\% & 1125\\ 
		\hline 
		3dmatch &   0.08 & 0.04 & 0.04 & 0.02 & 23\% & \bf 6.6\\ 		
		\hline 
		FDCP &    0.06 & \textbf{0.02} & 0.04 & \bf 0.01 & 40\% & 8.2\\ 
		\hline 
		Ours &    \textbf{0.04} &  0.03  &  0.04 & 0.02 & \bf 96\%  & 163\\ 
		\hline		
	\end{tabular} 	
\end{table} 

\subsection{Additional Experiments}
We have also done 14700 similar trials using 100 2D fish models from the Gatorbait100\footnote{http://www.rvg.ua.es/graphs/dataset01.html} database and have used the other 2D point clouds with 100 different shapes (face, umbrella, computer etc.) to test our sensitivity to shapes. We have used 100*5 2D models from Gatorbait100 database with 5 different densities to test our sensitivity to density. All the results are equally robust.  We have also tested our method without any uncertainty information and replaced each covariance matrix with an identity matrix. The results show our method with uncertainty information is better than that without uncertainty and both are better than the other comparison algorithms. For more details, see our supplementary materials \url{https://github.com/Canpu999/DUGMA}.

%
%

\section{Conclusion}
The proposed algorithm is simpler and more effective than the previous algorithms. We incorporated the 3D uncertainty distribution into a simple dynamic Gaussian mixture alignment.  The obvious difference between our algorithm and the previous ones is that it needs covariances at each point as input, which requires error models of how to estimate the real covariance for each kind of sensor.  All the experiments we have done show that the proposed method is very robust and accurate and works well. 
In the future, we will only use the points from set $\mathbf{Y}$ in the neighbourhood of ${x_i}$ to approximate all the points in set $\mathbf{Y}$ in Equation~\eqref{eq6} to reduce the time complexity to $\mathcal{O}(N)$.

 \section{ACKNOWLEDGMENTS}
The research is funded by TrimBot2020 project [Grant Agreement No. 688007] in  the European Union Horizon 2020 programme. We thank Dylan Campbell (\cite{GOGMA, SVGM}), Andy Zeng~\cite{3DMATCH}, Huan Lei~\cite{FDCP}, Jihua Zhu~\cite{BDICP} for providing help when we did the comparison experiments. We thank Dylan Campbell, Marcelo Saval-Calvo for giving us good advice on this paper.

{\small
\bibliographystyle{ieee}
\bibliography{egbib}
}

\end{document}